\PassOptionsToPackage{table}{xcolor}
\documentclass[acmtog]{acmart}
\acmSubmissionID{1411}

\usepackage{multirow}  % for spanning multiple rows
\usepackage{graphicx}  % for \rotatebox
\usepackage{makecell}
\newcommand{\rev}[1]{#1}

\usepackage[ruled]{algorithm2e} % For algorithms

\SetAlFnt{\small}
\SetAlCapFnt{\small}
\SetAlCapNameFnt{\small}
\SetAlCapHSkip{0pt}

\acmJournal{TOG}

\copyrightyear{2026}
\acmYear{2026}
\setcopyright{cc}
\setcctype{by}
\acmConference[SA Conference Papers '26]{SIGGRAPH Asia 2026 Conference Papers}{December 01--04, 2026}{Kuala Lumpur, Malaysia}
\acmBooktitle{SIGGRAPH Asia 2026 Conference Papers (SA Conference Papers '26), December 01--04, 2026, Kuala Lumpur, Malaysia}
\acmDOI{10.1145/3829340.3842214}
\acmISBN{979-8-4007-2842-6/2026/12}
\makeatletter
\DeclareRobustCommand\onedot{\futurelet\@let@token\@onedot}
\def\@onedot{\ifx\@let@token.\else.\null\fi\xspace}
 
\def\ie{i.e\onedot}

\makeatother

\begin{document}
% Title portion
\title{ADELE - Adaptive Delaunay Grids for High-Fidelity Mesh-Native Reconstruction}

% DO NOT ENTER AUTHOR INFORMATION FOR ANONYMOUS TECHNICAL PAPER SUBMISSIONS TO SIGGRAPH 2019!
\author{Johannes Weidenfeller}
\orcid{0000-0003-4845-0476}
\affiliation{%
 \institution{ETH Zurich}
 \city{Zurich}
 \country{Switzerland}}
\email{johannes.weidenfeller@ai.ethz.ch}

\author{Shaofei Wang}
\orcid{0000-0002-2865-698X}
\affiliation{%
 \institution{Beijing Institute for General Artificial Intelligence}
 \city{Beijing}
 \country{China}
}
\email{sfwang0928@gmail.com}

\author{Philipp Fürnstahl}
\orcid{0000-0001-6484-6206}
\affiliation{%
 \institution{University of Zurich}
 \city{Zurich}
 \country{Switzerland}
}
\affiliation{%
 \institution{Balgrist University Hospital}
 \city{Zurich}
 \country{Switzerland}
}

\author{Siyu Tang}
\orcid{0000-0002-1015-4770}
\affiliation{%
 \institution{ETH Zurich}
 \city{Zurich}
 \country{Switzerland}
}

%\author{Aparna Patel}
%\affiliation{%
% \institution{Rajiv Gandhi University}
% \streetaddress{Rono-Hills}
% \city{Doimukh}
% \state{Arunachal Pradesh}
% \country{India}}
%\email{aprna_patel@rguhs.ac.in}
%\author{Huifen Chan}
%\affiliation{%
%  \institution{Tsinghua University}
%  \streetaddress{30 Shuangqing Rd}
%  \city{Haidian Qu}
%  \state{Beijing Shi}
%  \country{China}
%}
%\email{chan0345@tsinghua.edu.cn}
%\author{Ting Yan}
%\affiliation{%
%  \institution{Eaton Innovation Center}
%  \city{Prague}
%  \country{Czech Republic}}
%\email{yanting02@gmail.com}
%\author{Tian He}
%\affiliation{%
%  \institution{University of Virginia}
%  \department{School of Engineering}
%  \city{Charlottesville}
%  \state{VA}
%  \postcode{22903}
%  \country{USA}
%}
%\affiliation{%
%  \institution{University of Minnesota}
%  \country{USA}}
%\email{tinghe@uva.edu}
%\author{Chengdu Huang}
%\author{John A. Stankovic}
%\author{Tarek F. Abdelzaher}
%\affiliation{%
%  \institution{University of Virginia}
%  \department{School of Engineering}
%  \city{Charlottesville}
%  \state{VA}
%  \postcode{22903}
%  \country{USA}
%}

%\renewcommand\shortauthors{Zhou, G. et al}

\begin{abstract}
Meshes remain the most practical representation for geometry reasoning and integration into graphics pipelines, yet existing reconstruction methods struggle to produce high-quality meshes. Most state-of-the-art approaches initially learn an intermediate representation (NeRF/3DGS) and treat mesh extraction as a post-processing step, which often leads to oversmoothed surfaces or poor quality meshes with excessive triangle counts.
Existing mesh-native optimization methods alleviate some of these issues but suffer from 
%poor scalability 
fixed-resolution discretizations and unstable optimization behavior. %The former limits the application scope to small-scale, object-centric scenes, while the latter leads to overall inferior reconstruction quality compared to radiance-field-based approaches.
% , and failing to capture high-frequency geometry.
In this paper, we introduce an adaptive mesh-based optimization framework and a practical mesh rendering technique to address these challenges. Our representation combines an optimizable Delaunay-triangulated tetrahedral grid with a multi-resolution hash grid. The former is refined through point pruning and insertion, while the latter provides latent features for SDF/appearance value predictions.
We use volumetric rendering to bootstrap a coarse geometry while leveraging mesh-based rendering for recovering fine-grained details. Additionally, we propose a differentiable, rasterization-based depth-offset rendering formulation, reducing geometric artifacts and improving reconstruction quality.
% Our method achieves state-of-the-art results on DTU and Tanks and Temples, quantitatively demonstrating superior recovery of high-curvature geometry while remaining fully compatible with mesh regularization terms that further enhance mesh quality.
Our method significantly outperforms existing mesh optimization approaches across a variety of object-centric benchmarks while being competitive with state-of-the-art NeRF/3DGS methods. Code and additional results are available on our
\href{https://johannes-weidenfeller.github.io/adele}{project page}.
\end{abstract}

%
% The code below should be generated by the tool at
% http://dl.acm.org/ccs.cfm
% Please copy and paste the code instead of the example below.
%


%
% End generated code
%


\begin{teaserfigure}
   \includegraphics[width=\linewidth]{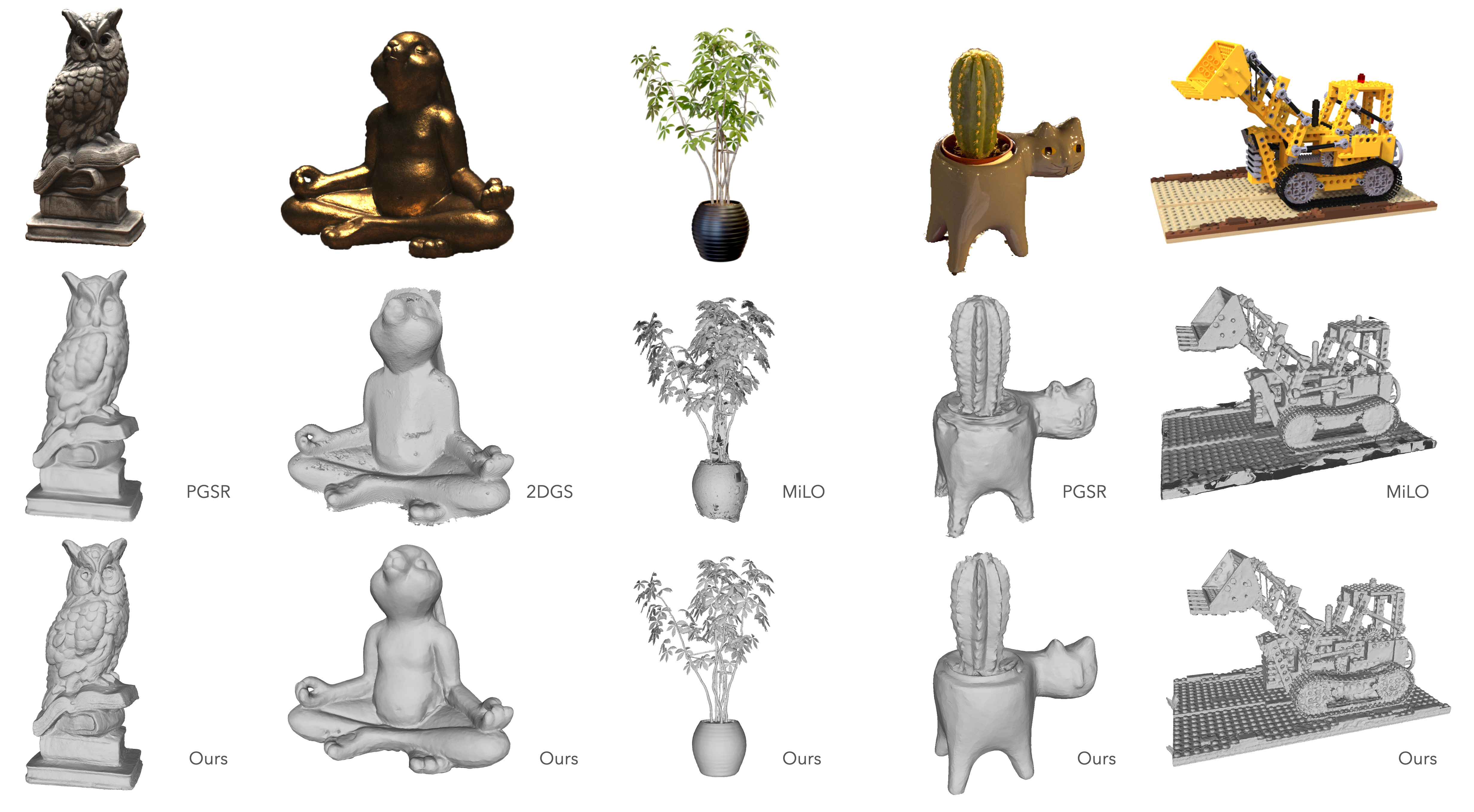}
    \caption{Qualitative comparison of mesh reconstructions from multi-view images. Existing methods based on 3D Gaussian Splatting (PGSR, 2DGS) or mesh-gaussian-hybrid optimization (MiLO) produce meshes with noticeable artifacts or  oversmoothed surfaces. Our method directly optimizes a locally refinable tetrahedral grid with a novel depth-offset rendering formulation, yielding high-quality meshes with sharp geometric details and well-distributed triangles.}
\end{teaserfigure}
  
\maketitle

\section{Introduction}
\label{sec:intro}
Reconstructing detailed geometry from multiview images is a long-standing problem in computer vision and graphics, with numerous applications in 3D modeling, virtual reality, and robotics. Recent advances in neural radiance fields (NeRF)~\cite{Mildenhall2020ECCV} and 3D Gaussian Splatting (3DGS)~\cite{Kerbl2023TOG} have led to a blossoming of radiance field-based reconstruction methods~\cite{Wang2021NEURIPS,Yariv2021NEURIPS,Oechsle2021ICCV,Li2023CVPR,Guedon2024CVPR,Huang2024SIGGRAPH,Yu2024TOG}, which achieve unprecedented levels of detailed geometry recovery from only a small set of images
(e.g., 20-100 images). However, these methods rely on either neural fields or Gaussians to represent the radiance fields, which are less suitable for standard graphics pipelines, where polygonal meshes remain the most practical representation for geometry reasoning and downstream tasks.
%However, these methods rely on either neural fields or Gaussians to represent the radiance fields; these representations are not directly compatible with standard graphics pipelines, where polygonal meshes remain the most practical representation for geometry reasoning and downstream tasks. 
Consequently, most existing works treat mesh extraction as a post-processing step, typically using Marching Cubes/Tetrahedra~\cite{Wang2021NEURIPS,Yariv2021NEURIPS,Oechsle2021ICCV,Li2023CVPR,Yu2024TOG}, TSDF fusion~\cite{Huang2024SIGGRAPH,Guedon2025CVPR}, or Poisson surface reconstruction~\cite{Peng2021NEURIPS,Guedon2024CVPR,Huang2024SIGGRAPH,Guedon2025CVPR}. This often leads to either a loss of fine geometric details or excessively dense meshes that are impractical for real-world applications. Furthermore, the extracted meshes usually exhibit reduced rendering fidelity compared to the intermediate representation.

Another line of work incorporates the mesh extraction step in the optimization pipeline and optimizes meshes directly in an end-to-end fashion~\cite{Munkberg2022CVPR,Hasselgren2022NEURIPS,Fruehauf2024CVPR}. These methods provide a direct link between the optimized mesh and the rendered output and allow the use of regularizers to improve mesh quality. 
However, existing approaches typically rely on fixed-resolution discretizations, which limit the adaptive allocation of geometric detail and constrain reconstruction fidelity, particularly in regions with complex geometry. In addition, optimizing explicit mesh representations remains sensitive to local geometric artifacts and unstable surface updates during training.
%While conceptually appealing, the discontinuous nature of meshes makes them hard to optimize, and reconstructed meshes from existing methods often suffer from significant artifacts. Furthermore, these approaches assume uniform grids, which constrain the reconstruction resolution and inherently limit the fidelity of the recovered geometry.

% More recently, works have focused on incorporating extracted meshes into the reconstruction pipeline in a hybrid fashion. 

% We propose a novel end-to-end mesh optimization algorithm that addresses the shortcomings of existing reconstruction methods. To improve scalability, our representation is built on an optimizable grid that is periodically rebuilt using Delaunay triangulation. The grid can be initialized from a coarse geometric estimate derived from VGGT depth maps, and adaptively refined during optimization through point pruning and insertion. This allows our method to extract meshes with fine geometric detail and locally adaptive resolution while remaining computationally efficient. To enhance the stability of the optimization, we employ a differentiable mesh rendering strategy that improves gradient flow through a lightweight depth-offset sampling scheme. Since the optimization is performed in an end-to-end fashion, regularization terms that improve mesh quality can be seamlessly integrated into the training objective.
In this paper, we present \textbf{ADELE}, an adaptive mesh-native reconstruction framework designed to address the limitations of existing multi-view mesh reconstruction methods.
%In this paper, we propose a new mesh-based optimization algorithm that addresses the shortcomings of existing multi-view reconstruction methods. 
To enable high-resolution reconstruction, our representation is built on an optimizable tetrahedral grid that is periodically rebuilt using Delaunay triangulation~\cite{Govindarajan2025ICCV,Guedon2025TOG,Binninger2025TOG}. The grid can be initialized uniformly and adaptively refined during optimization through point pruning and insertion. This allows us to extract meshes with fine geometric detail and locally adaptive resolution while remaining computationally efficient. To enhance optimization stability, we utilize a hybrid approach combining volumetric and mesh-based rendering. Specifically, we found that using volumetric rendering in early training stages stabilizes training, but cannot produce sharp details if employed in later stages; we thus use volumetric rendering to bootstrap a coarse geometry and later transition to mesh-based rendering to recover fine-grained details. Additionally, we employ a differentiable mesh rendering strategy based on a lightweight depth-offset sampling scheme. Since the optimization is carried out in an end-to-end fashion, regularization terms that improve mesh quality can be seamlessly integrated into the training objective.

% To summarize, our contributions are as follows.
% \begin{itemize}
% \item{We propose a hybrid geometric representation that combines an adaptive grid-based representation using Delaunay triangulation with a multi-resolution hash encoding, enabling flexible geometry learning from random/uniform initialization.}
% \item{We introduce a differentiable mesh rendering strategy, sampling colors at multiple depth offsets around the rasterized surface.}
% % \item{We combine an adaptive grid-based representation using Delaunay triangulation with a pruning and densification strategy that allows for local adaptation to highly detailed scene geometry while remaining computationally feasible.}
% \item{We demonstrate that our method is compatible with regularization terms that improve the quality of the extracted meshes, making them more suitable for downstream tasks.}
% \item We validate our method through extensive evaluations on various datasets (DTU, BlendedMVS, NeRF Synthetic and Stanford ORB), significantly outperforming state-of-the-art mesh-based reconstruction methods in terms of geometric reconstruction. We also provide evidence that our method can achieve detailed geometry reconstruction on unbounded, large-scale scenes such as the Tanks and Temple dataset.
% \end{itemize}
To summarize, our contributions are as follows.
\begin{itemize}
\item{We propose a hybrid geometric representation that combines an adaptive grid-based representation using Delaunay triangulation with a multi-resolution hash encoding, enabling flexible geometry learning from random or uniform initialization. The representation supports hybrid optimization using both volumetric and mesh-based rendering supervision, substantially improving optimization stability during early training.}

\item{We introduce a differentiable mesh rendering strategy that samples colors at multiple depth offsets around the rasterized surface.}

% \item{We demonstrate that our method is compatible with mesh regularization terms that improve the quality of the extracted meshes, making them more suitable for downstream applications.}

\item{We analyze biases in the DTU evaluation protocol, showing that it can favor mesh extraction strategies with open or incomplete surfaces. To mitigate this issue, we propose a simple filtering scheme that enables more consistent geometric evaluations across reconstruction methods.}

\item{We validate our method through extensive experiments on DTU, BlendedMVS, NeRF Synthetic, and Stanford ORB, significantly outperforming state-of-the-art mesh-based reconstruction methods in terms of geometric reconstruction quality. }
%We additionally demonstrate geometry reconstruction on large-scale unbounded scenes from the Tanks and Temples dataset.}
\end{itemize}

% \item{We propose a novel grid-based scene representation that extends a signed distance function to a local Gaussian distribution, allowing us to leverage stochastic rendering for optimizing the underlying representation.}
\section{Related Work}
\label{sec:related_work}

\subsection{Radiance Fields}
The introduction of Neural radiance fields (NeRF)~\cite{Mildenhall2020ECCV} led to major advancements in the task of novel view synthesis. Follow-up works improved its anti-aliasing abilities~\cite{Barron2021ICCV,Hu2023ICCV}, generalization and scalability to large, unbounded environments~\cite{Barron2022CVPR}, and explored more efficient representations using voxel grids~\cite{Yu2022CVPR,Sun2022CVPR,Mueller2022TOG} or tensor decompositions~\cite{Chen2022ECCV,Chen2023ARXIVb,Chen2023TOG}. More recently, 3D Gaussian Splatting~\cite{Kerbl2023TOG} (3DGS) has enabled lightning fast optimization and real-time rendering of 3D scenes with impressive fidelity, on both object-centric and unbounded scenes. Since then, the community has witnessed a surge of interest in this representation, and an abundance of follow-up works were proposed to further improve its efficiency~\cite{Fang2024ECCV,Fang2024ARXIV,Ye2024JMLR,Kheradmand2025ARXIV} and rendering quality~\cite{Yu2024CVPR,Liang2024ECCV}.

\subsection{Neural Surface Reconstruction}
A number of works~\cite{Yariv2021NEURIPS,Wang2021NEURIPS} extend NeRF for surface reconstruction by converting learnable signed distance fields (SDFs) into volumetric density, followed by volumetric rendering. Further advances~\cite{Alexandru2023CVPR,Li2023CVPR,Wang2023ICCV} improve the efficiency and extend applications to large-scale scenes by leveraging efficient multi-resolution feature grids~\cite{Mueller2022TOG}. 
More recently, several works have shifted focus toward surface reconstruction through 3DGS. Early approaches improve the surface quality extracted from 3DGS scenes by introducing explicit regularization terms~\cite{Guedon2024CVPR}, or by training hybrid representations that jointly optimize a 3DGS scene and an SDF~\cite{Chen2023ARXIVa,Yu2024NEURIPS}. Methods such as 2DGS~\cite{Huang2024SIGGRAPH} and GaussianSurfels~\cite{Dai2024SIGGRAPH} model scenes with 2D Gaussian primitives that more naturally align with the underlying surface geometry. % and additionally enforce depth–normal consistency to enhance geometric accuracy.
Despite these advances, most Gaussian-based reconstruction methods still rely on surface extraction performed after training, typically using TSDF fusion or Poisson surface reconstruction, which often produce oversmoothed surfaces and fail to capture high-curvature details. To overcome this limitation, Gaussian Opacity Fields (GOF)~\cite{Yu2024TOG} propose building a tetrahedral grid from the bounding boxes of Gaussians and extracting iso-surfaces from these tetrahedral grids. However, this mesh extraction strategy remains a separate post-processing step, leading to a mismatch in the optimized representation and the extracted surface. 

\subsection{Mesh-based Reconstruction}
Mesh-based reconstruction methods directly optimize mesh geometry through differentiable rendering. Early work optimized vertex positions under a fixed initial topology or relied on heuristics for topology changes~\cite{Chen2019NEURIPS,Chen2021NEURIPS,Jatavallabhula2019ARXIV}. NVDiffrec~\cite{Munkberg2022CVPR} jointly optimizes geometry, texture, and lighting on a uniform tetrahedral-grid SDF, with follow-ups improving mesh extraction~\cite{Shen2023TOG,Fruehauf2024CVPR} or adopting a more physically plausible rendering model~\cite{Hasselgren2022NEURIPS}. These methods, however, remain bottlenecked by uniform grid resolution and cannot recover fine geometry.
Adaptive Delaunay and tetrahedral structures~\cite{delaunay1934sphere,lee1980two} have been explored for radiance-field~\cite{Kulhanek2023ICCV,Gu2024NEURIPS} and, more recently, mesh-based reconstruction. TetWeave~\cite{Binninger2025TOG} introduces an adaptive Delaunay grid for efficient mesh extraction; its extraction is conceptually close to ours, but its multi-view reconstruction is restricted to object-centric scenes with ground-truth depth supervision.
RadiantFoam~\cite{Govindarajan2025ICCV} renders directly with Voronoi cells (the dual of Delaunay), but targets novel view synthesis only and does not extract meshes.
\rev{Triangle Splatting~\cite{Held_2026_3DV} and MeshSplatting~\cite{Held_2026_CVPR}
optimize unstructured triangle soups for high-quality rendering of large-scale
scenes, with the latter connecting the primitives into a mesh through a single
Delaunay triangulation at the end of optimization. The resulting meshes are well suited for rendering, but exhibit irregularities and limited accuracy as
surface reconstructions.} 
MILo~\cite{Guedon2025TOG} extracts meshes from Delaunay sites derived from Gaussian bounding boxes, yet renders through the Gaussian branch and supervises the mesh only via mutual depth/normal losses rather than image gradients. Our method instead optimizes an adaptive Delaunay tetrahedral grid end-to-end with image-space losses,
enabling local refinement of fine-grained detail while remaining computationally efficient.

\subsection{Mesh-based Rendering}

Another central challenge in mesh-based reconstruction lies in the discontinuous nature of mesh rendering, which complicates optimization. Differentiable renderers help mitigate this problem by enabling gradient flow through discontinuities. NVDiffrast~\cite{Laine2020TOG} introduces edge anti-aliasing to provide gradients along triangle boundaries, while alpha-blending approaches~\cite{Ravi2020ARXIV,Liu2019ICCV} propagate gradients across occluded surfaces at the cost of rendering efficiency. Another line of work leverages meshes as an auxiliary data structure to accelerate rendering and improve appearance quality. To approximate volumetric samples  and better handle fuzzy geometry, several approaches construct multi-layer structures around an extracted base mesh~\cite{adaptiveshells2023, Esposito2025VolSurfs} or bake trained radiance fields into optimized polygon layers for efficient inference~\cite{quadfields}. Similarly, other methods~\cite{Guedon2024CVPR, guedon2024frosting} rely on a mesh to anchor surface-aligned 3D Gaussians or to confine them within an enclosing shell. Binary Opacity Grids~\cite{Reiser2024SIGGRAPH} instead anneal a sparse voxel grid of soft opacities into hard occupancies before mesh extraction. However, these techniques are predominantly evaluated on novel view synthesis, with little to no analysis of the quality of the recovered surface itself. Most closely related is the concurrent work \rev{Mesh Splatting}~\cite{zhang2026mesh}, which softens a mesh into semi-transparent blended layers. While conceptually similar to our method, it exhibits lower performance on the evaluated datasets and struggles to capture high-frequency details. Our approach instead builds directly on NVDiffrast, retaining its computational efficiency while introducing a locally adaptive sampling mechanism that, as shown in our ablations, stabilizes training and reduces geometric artifacts.

\section{Method}
\label{sec:method}

\begin{figure*}[t]
  \centering
  \includegraphics[width=1.0\linewidth]{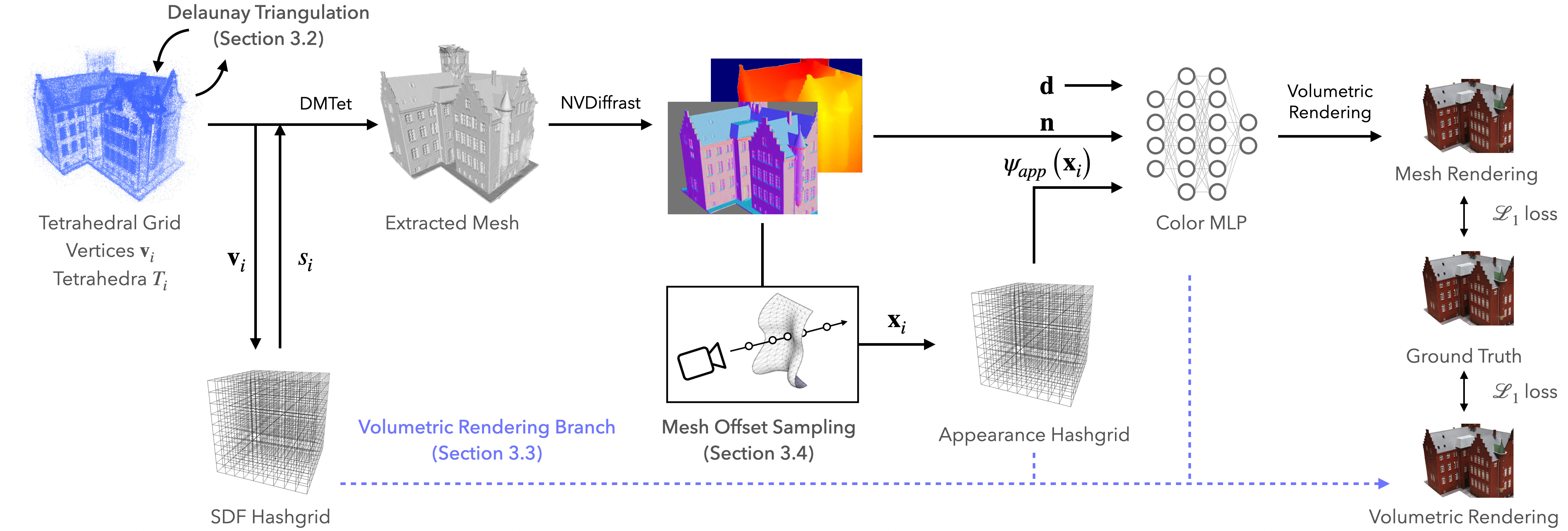}
    \caption{\textbf{Overview of our framework.} We represent scene geometry using an adaptive tetrahedral grid whose connectivity is maintained via Delaunay triangulation (\S~\ref{subsec:geometric_representation}) and whose vertex positions are optimizable (\S~\ref{subsec:grid_optimization}). SDF values at grid vertices are queried from a multi-resolution hash grid, and a mesh is extracted using the Marching Tetrahedra algorithm (DMTet). The mesh is rasterized via NVDiffrast to obtain depth and normal maps. For each pixel, we perform depth-offset sampling (\S~\ref{subsec:differentiable_mesh_rendering}) to generate multiple 3D sample points $\mathbf{x}_i$ around the surface. These points, along with the viewing direction $\mathbf{d}$ and surface normal $\mathbf{n}$, are fed through an appearance hash grid and a color MLP to produce pixel colors. During early training, a volumetric rendering branch (dashed blue) provides additional supervision to stabilize optimization (\S~\ref{subsec:hybrid_training}). The final output is supervised with an $\mathcal{L}_1$ photometric loss against ground-truth images along with mask and mesh regularization losses (\S~\ref{subsec:regularization_and_loss}).}
   \label{fig:pipeline}
\end{figure*}

% In this section, we describe our method, as shown in Fig.~\ref{fig:pipeline}
\subsection{Geometric Representation}
\label{subsec:geometric_representation}

%TODO: This section still misses a description of the background model and the applied space contraction for large scale scenes
We represent the scene geometry using a discretized signed distance function (SDF), combining a tetrahedral grid with a multi-resolution hash grid. Formally, the tetrahedral grid is defined by a set of \(n_v\) vertices \(\mathbf{V}\) and \(n_t\) tetrahedra \(\mathbf{T}\):
\[
\mathbf{V} = \{ \mathbf{v}_i \in \mathbb{R}^3 \mid i = 1, \dots, n_v \}, \quad
\mathbf{T} \subseteq \mathbf{V}^4.
\]

We encode the SDF using a multi-resolution hash grid \(\hat{s}: \mathbb{R}^3 \to \mathbb{R}\). To extract a mesh from this representation, we query the SDF at the vertices of the tetrahedral grid to obtain discrete SDF samples
\(s_i = \hat{s}(\mathbf{v}_i)\).
We extract a mesh by applying the Marching Tetrahedra algorithm~\cite{marching_tetrahedra}.
%, an adaptation of Marching Cubes for tetrahedral grids.
Specifically, for any edge connecting vertices \((\mathbf{v}_i, \mathbf{v}_j)\) with differing SDF signs,
%\(\text{sign}(s_i) \neq \text{sign}(s_j)\),
a mesh vertex \(\mathbf{v}_{ij}\) is placed along the edge according to
\[
\mathbf{v}_{ij} = \frac{\lvert s_i \rvert\, \mathbf{v}_j + \lvert s_j \rvert\, \mathbf{v}_i}{\lvert s_i \rvert + \lvert s_j \rvert}.
\]
Finally, the mesh vertices extracted from any tetrahedron are connected with one or two triangle faces, resulting in a watertight mesh.

\subsection{Grid Optimization}
\label{subsec:grid_optimization}

% \begin{figure}[h]
%   \centering
%    \includegraphics[width=0.9\linewidth]{figures/adaptive_grid.jpg}

%    \caption{\textbf{Adaptive tetrahedral grid optimization.} Starting from a uniform initialization (left), the grid progressively adapts to the scene geometry through Delaunay-site (\ie, vertex) movement, densification, and pruning. Periodic Delaunay re-triangulation maintains valid grid connectivity throughout optimization. The rightmost column shows the final extracted meshes.}
%    \label{fig:grid_optimization}
% \end{figure}

In our method, we treat the vertex positions of the tetrahedral grid as optimizable parameters.
% To maintain a valid grid structure and avoid intersecting tetrahedra during the course of the optimization,
Inspired by~\cite{Govindarajan2025ICCV}, we periodically update the grid connectivity using Delaunay triangulation. This strategy also enables the addition and removal of grid vertices, analogous to the densification and pruning used in 3DGS. New vertices are inserted at fixed iterations, ensuring that the grid progressively captures finer scene details. Similar to common densification schemes in other representations~\cite{Kerbl2023TOG, Govindarajan2025ICCV}, we use a gradient-based heuristic to identify regions requiring higher resolution. For each grid vertex $\mathbf{v}_i$, we maintain an exponential moving average of the positional gradient magnitude and periodically sample from existing tetrahedrons for subdivision; the probability of selecting a tetrahedron $T_j$ is proportional to the average tracked gradient magnitude of its corner vertices $\mathbf{v}_{j_1}, \dots, \mathbf{v}_{j_4}$:
\[
p(T_j) \propto \frac{1}{4} \sum_{k=1}^{4} \overline{|\nabla_{\mathbf{v}_{j_k}} \mathcal{L}|}.
\]
Here $\overline{|\nabla_{\mathbf{v}_{j_k}} \mathcal{L}|}$ denotes the exponential moving average of the positional gradient magnitude at vertex $\mathbf{v}_{j_k}$. This sampling strategy ensures that densification focuses on regions contributing the most to the optimization error. For a sampled tetrahedron, a new vertex \(\mathbf{v}_{\text{new}}\) is inserted by sampling barycentric coordinates \(b_i\) from a uniform Dirichlet distribution:
\begin{align}
    \mathbf{v}_{\text{new}} = \sum_{k=1}^{4} b_k \, \mathbf{v}_{j_k}, \qquad \text{with} \sum_{i=1}^{4} b_i = 1.
\end{align}

% The base signed distance value of the new grid point, \(s^0_{\text{new}}\), is initialized as follows
% \[
% s^0_{\text{new}} = \sum_{k=1}^{4} b_k \, s_{j_k} - \hat{s}\left( \mathbf{v}_\text{new} \right),
% \]
% ensuring that the effective SDF value of the added vertex matches the corresponding barycentric interpolation of the SDF values at the tetrahedron corners. 

\begin{figure}[h]
  \centering
   \includegraphics[width=\linewidth]{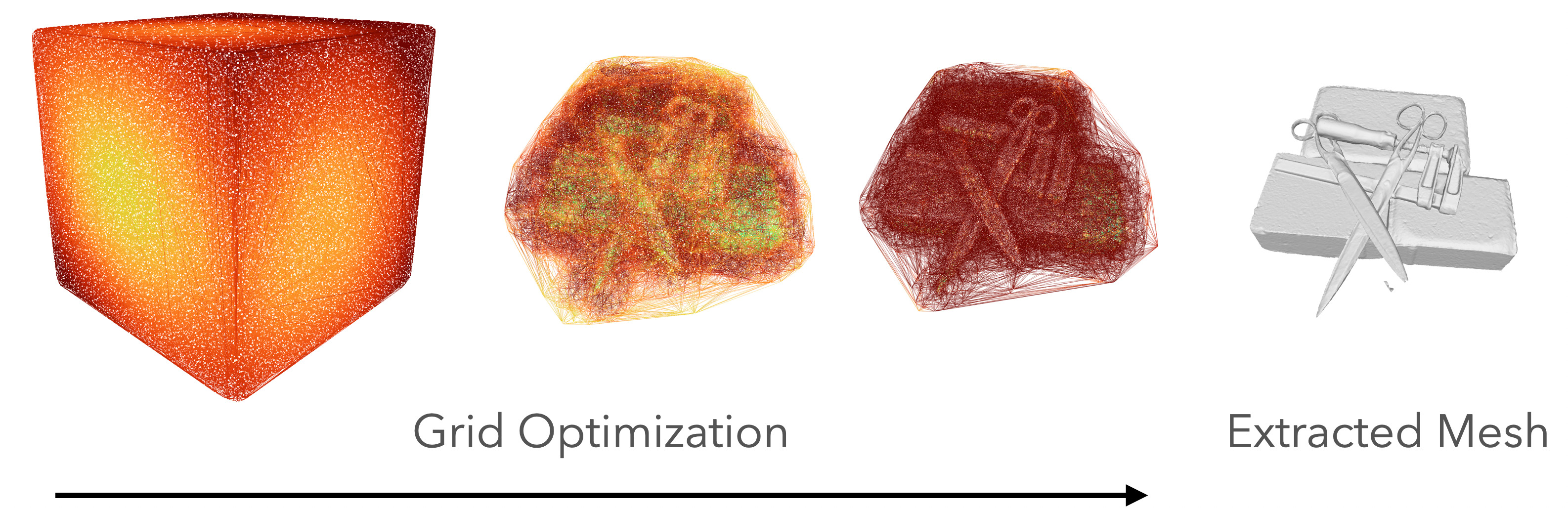}

   \caption{\textbf{Adaptive tetrahedral grid optimization.} Starting from a uniform initialization (left), the grid progressively adapts to the scene geometry through Delaunay-site (\ie, vertex) movement, densification, and pruning. Periodic Delaunay re-triangulation maintains valid grid connectivity throughout optimization.}
   \label{fig:grid_optimization}
\end{figure}

To maintain computational efficiency and eliminate unnecessary interior geometry, we also regularly prune vertices based on their contribution to the scene. We define a vertex as \textbf{passive} if it does not contribute to the rendered output of any training view. We periodically remove these passive vertices, retaining only those that are immediate neighbors of active (contributing) vertices to maintain a valid support region. Figure \ref{fig:grid_optimization} shows how the grid adapts to the reconstructed geometry as the optimization progresses. The combination of adaptive densification and pruning ensures that the tetrahedral grid remains both expressive and computationally tractable throughout the optimization process.

\subsection{Hybrid Training}
\label{subsec:hybrid_training}

We find that optimizing our representation using purely mesh-based inverse rendering is highly sensitive to the initial state of the SDF. To enhance optimization stability, we introduce a hybrid training strategy. Specifically, we utilize a volumetric rendering objective to bootstrap a stable, coarse geometry during the early stages of training. As the optimization progresses, we smoothly transition to mesh-based rendering, which is better suited for recovering sharp, fine-grained surface details.

\paragraph{Mesh-based rendering}
For a given input view, we rasterize the mesh extracted from the tetrahedral grid using NVDiffrast~\cite{Laine2020TOG} to obtain anti-aliased normal and depth maps. 
Let \(d\) and \(\mathbf{n}\) denote the depth and normal of a pixel, respectively, and let $\mathbf{x}$ be the 3D point obtained by unprojecting $d$. We predict the color $c$ using a shallow network that takes as input an appearance feature of the 3D location $\mathbf{x}$, the encoded viewing direction $\mathbf{d}$, and the surface normal $\mathbf{n}$:
\begin{equation}
\label{eqn:color_network}
c = \mathrm{MLP}_{col}\Big( \Phi_{app}\left( \mathbf{x} \right),  \mathbf{d}, \Phi_{n}\left(\mathbf{n}\right) \Big),
\end{equation}
where $\Phi_{app}$ is a multi-resolution hash encoding and $\Phi_{n}$ is a spherical harmonic encoding.

\paragraph{Volumetric Rendering}
Leveraging the underlying hash grid encoding, we adopt the formulation of NeuS~\cite{Wang2021NEURIPS} to transform SDF values into volumetric densities and compute compositing weights for samples along each ray. The color at each sample location is evaluated via \autoref{eqn:color_network} where the surface normals are derived from the analytical gradient of the hash-grid-defined SDF. We phase out the volumetric supervision by gradually reducing its loss weight over the course of training.
%This ensures a smooth transition where the optimization begins with volumetric guidance but concludes using the mesh-based branch exclusively.
Using only volumetric rendering for the whole training process is detrimental to performance, as shown in our experiments (Sec.~\ref{subsec:ablation_studies}).
 % ; we hypothesize that this is due to volumetric sampling of NeuS and similar methods generating random samples \emph{near} the surface but not \emph{on} the surface, resulting in smooth reconstructions which lack details.
Further details regarding the volumetric optimization branch are provided in the supplementary material. % \ref{sec:A_volumetric_rendering_details}.

\subsection{Depth Offset Sampling}
\label{subsec:differentiable_mesh_rendering}

% \begin{figure}[h]
%   \centering
%    \includegraphics[width=\linewidth]{figures/mesh_offset_sampling2.jpg}
%     \caption{\textbf{Depth-offset sampling.} (Left) We sample multiple points along each camera ray by performing inverse sampling with respect to the opacity function $\Psi(t)$. The sigmoid shape of $\Psi$ concentrates samples near the mesh surface, where the SDF transitions through zero. (Right) early stage reconstruction rendered using sampled depth offsets. }
%    \label{fig:bandwidth_sampling}
% \end{figure}

\begin{figure}[h]
  \centering
   \includegraphics[width=\linewidth]{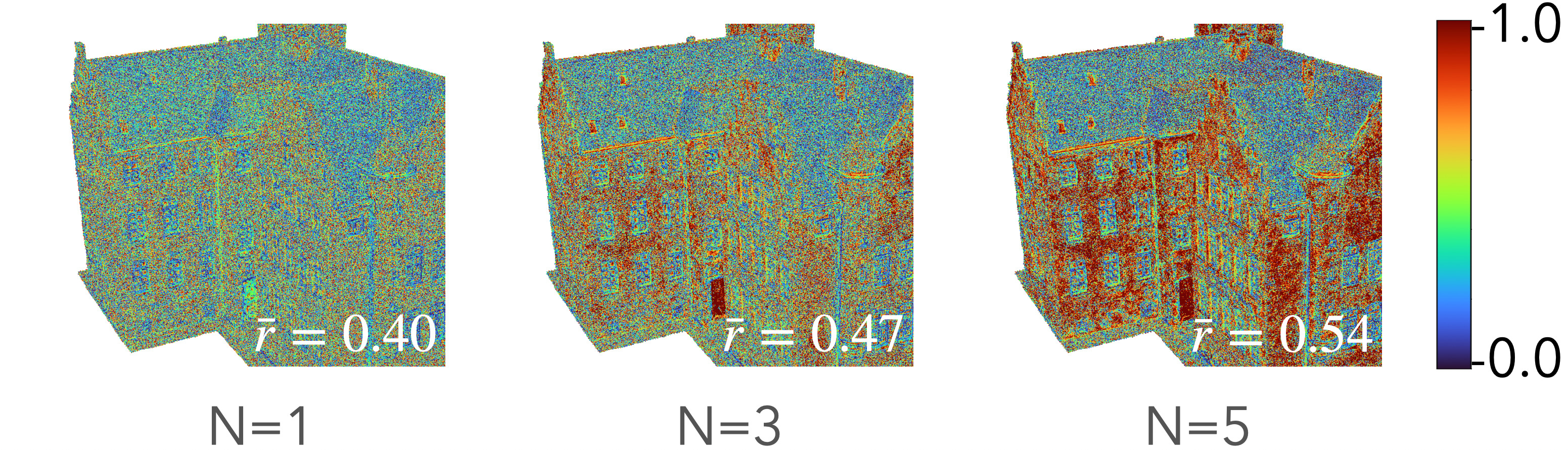}
    \caption{\textbf{Depth-offset sampling.} Per-pixel alignment between the RGB reconstruction loss gradient w.r.t. mesh position and the viewing direction for different numbers of offset samples $N$. Red indicates strong alignment with the viewing direction, while blue denotes largely perpendicular gradients. Increasing $N$ leads to progressively stronger alignment with the viewing direction.}
   \label{fig:gradient_analysis}
\end{figure}

Despite the hybrid supervision signal, the optimization at times becomes trapped in local minima, leading to undesirable convex artifacts (see Fig.~\ref{fig:ablation_results}). To facilitate training via the mesh-based rendering path, we introduce a differentiable depth-offset rendering strategy. Rather than evaluating a single point, we sample multiple depth values $d_i$ around the rendered depth $d$ for each pixel and blend their corresponding colors. 
%The extent of this sampling region is governed by a jointly optimized bandwidth parameter, with exact sampling details provided in the supplementary material. 
We model the SDF as locally planar in orientation of the mesh normal. Similar to NeuS \cite{Wang2021NEURIPS}, we assume that the opacity along the ray follows a sigmoid curve centered at the intersection point, that is,
\[
\Psi(t) = \frac{1}{1+e^{\sigma f(t)}},
\]
where $f(t)$ denotes the SDF along the ray parameterized by $t$. Here, $\sigma$ is a learnable parameter that controls the steepness of the opacity function. To draw $n$ depth samples, we perform inverse sampling with respect to $\Psi$. For each sampled position $\mathbf{x}_i$, its corresponding color $c_i$ is computed via Eq.~\ref{eqn:color_network}. The final pixel color is then obtained as a weighted average over all sampled colors according to their contribution $w_i$ to the ray opacity
\begin{equation}
C = \sum_i w_i c_i.
\end{equation}
Details of the sampling procedure and the computation of the weights $w_i$ are included in the supplementary material.
Since we perform inverse sampling with respect to the opacity function, the parameter $\sigma$ controls the \rev{width} of the sampling region around the extracted mesh. In practice, we optimize this parameter as part of the training.

To analyze the impact of this strategy on the optimization dynamics, we evaluate how depth-offset rendering affects the alignment of mesh gradients $\mathbf{g}$ with the viewing direction $\mathbf{d}$ by measuring their cosine similarity $r = |\mathbf{g}^{\top}\mathbf{d}| / \|\mathbf{g}\|$. Figure~\ref{fig:gradient_analysis} illustrates $r$ for $N=1$, $3$, and $5$ mesh offset samples. The results show that increasing the sample count yields gradients that are more strongly oriented along the viewing direction. By encouraging updates that shift the surface perpendicular to the image plane, the optimization can more effectively escape local minima, facilitating the correction of convex artifacts and the recovery of concave structures.

\subsection{Training Losses}
\label{subsec:regularization_and_loss}
We directly supervise our representation using a photometric $L_1$-loss between the input images and the images rendered from our model, 
\begin{equation}
\label{eq:photometric_loss}
\mathcal{L}_{\text{photo}} = \sum_{c \in C} \left\| I_c - \hat{I}_c \right\|_1,
\end{equation}
where $I_c$ denotes the input image for camera $c$, $\hat{I}_c$ denotes the corresponding rendered image, and $C$ is the set of all cameras. When available, we also employ a mask loss to supervise the rendered opacity, given by
\begin{equation}
\label{eq:mask_loss}
\mathcal{L}_{\text{mask}} = \sum_{c \in C} \left\| M_c - \hat{M}_c \right\|_2^2,
\end{equation}
where $M_c$ denotes the ground-truth mask for camera $c$ and $\hat{M}_c$ denotes the rendered opacity mask. Our full loss function is then given by
\[
\mathcal{L} =  \lambda_\text{photo} \mathcal{L}_{\text{photo}} + \lambda_{\text{mask}} \mathcal{L}_{\text{mask}},
\]
%where $\lambda_\text{photo}$ and $\lambda_{\text{mask}}$ are weighting factors that balance the contribution of the individual loss terms. We additionally implement a triangle regularization loss (see \autoref{sec:D_mesh_regularization}), but find that it slightly degrades geometric accuracy and therefore do not use it in our main method.
\section{Experiments}
\label{sec:experiments}

\begin{table*}[h]
\caption{\textbf{Quantitative comparison of geometric reconstruction on the DTU dataset.} We report Chamfer distance (mm, $\downarrow$) for 15 test scenes, grouped by representation type: implicit (NeRF-based), Gaussian (3DGS-based), and mesh-based methods. Colors indicate \colorbox{red!30}{1st}, \colorbox{orange!40}{2nd}, and \colorbox{yellow!40}{3rd} place per scene. Our method achieves lower mean Chamfer distance among mesh-based approaches while performing on par with state-of-the-art radiance-field methods. Baseline runtimes are from~\cite{Guedon2025TOG} (RTX 4090, 24\,GB); ours were measured on an H200 (141\,GB)*. 
%A more extensive runtime and memory analysis is included in the supplementary material.
}
\label{tab:dtu_results}
\centering
\footnotesize
\renewcommand{\arraystretch}{1.0}
\setlength{\tabcolsep}{4pt}
% --- helper colors ---
% best     -> red
% second   -> orange
% third    -> yellow

\begin{tabular}{l l ccccccccccccccc c c}
\midrule
 & & 24 & 37 & 40 & 55 & 63 & 65 & 69 & 83 & 97 & 105 & 106 & 110 & 114 & 118 & 122 & \textbf{Mean} & \textbf{Time} \\
\cmidrule(lr){3-17} \cmidrule(lr){18-19}

% --- implicit section ---
\multirow{4}{*}{\rotatebox{90}{implicit}} &
NeRF~\cite{Mildenhall2020ECCV}
& 1.90 & 1.60 & 1.85 & 0.58 & 2.28 & 1.27 & 1.47 & 1.67 & 2.05 & 1.07 & 0.88 & 2.53 & 1.06 & 1.15 & 0.96 & 1.49 & $>12h$ \\

& VolSDF~\cite{Yariv2021NEURIPS}
& 1.14 & 1.26 & 0.81 & 0.49 & 1.25 & 0.70 & 0.72 & 1.29 & 1.18 & 0.70 & 0.66 & 1.08 & 0.42 & 0.61 & 0.55 & 0.86 & $>12h$ \\

& NeuS~\cite{Wang2021NEURIPS}
& 1.00 & 1.37 & 0.93 & 0.43 & 1.10 & 0.65 & 0.57 & 1.48 & 1.09 & 0.83 & 0.52 & 1.20 & 0.35 & 0.49 & 0.54 & 0.84 & $>12h$ \\

& Neuralangelo~\cite{Li2023CVPR}
& \cellcolor{orange!40} 0.37
&  \cellcolor{yellow!40} 0.72
& 0.35
& 0.35
& 0.87
& \cellcolor{orange!40}0.54
& \cellcolor{yellow!40} 0.53
& 1.29
& 0.97
& 0.73
& \cellcolor{orange!40} 0.47
& 0.74
& \cellcolor{orange!40} 0.32
& 0.41
& 0.43
& \cellcolor{yellow!40} 0.61
& {$>12h$} \\

\cmidrule(lr){1-2} \cmidrule(lr){3-17} \cmidrule(lr){18-19}

% --- Gaussian section ---
\multirow{8}{*}{\rotatebox{90}{Gaussian}} &
3DGS~\cite{Kerbl2023TOG}
& 2.14 & 1.53 & 2.08 & 1.68 & 3.49 & 2.21 & 1.43 & 2.07 & 2.22 & 1.75 & 1.79 & 2.55 & 1.53 & 1.52 & 1.50 & 1.96 & 11.2 m \\

& SuGaR~\cite{Guedon2024CVPR}
& 1.47 & 1.33 & 1.13 & 0.61 & 2.25 & 1.71 & 1.15 & 1.63 & 1.62 & 1.07 & 0.79 & 2.45 & 0.98 & 0.88 & 0.79 & 1.33 & $\sim$ 1h \\

& GaussianSurfels~\cite{Dai2024SIGGRAPH}
& 0.66 & 0.93 & 0.54 & 0.41 & 1.06 & 1.14 & 0.85 & 1.29 & 1.53 & 0.79 & 0.82 & 1.58 & 0.45 & 0.66 & 0.53 & 0.88 & 6.7 m \\

& 2DGS~\cite{Huang2024SIGGRAPH}
& 0.48
& 0.91
& 0.39
& 0.39
& 1.01
& 0.83
& 0.81
& 1.36
& 1.27
& 0.76
& 0.70
& 1.40
& 0.40
& 0.76
& 0.52
& 0.80
& 10.9 m \\

& GOF~\cite{Yu2024TOG}
& 0.50 & 0.82 & 0.37 & 0.37 & 1.12 & 0.74 & 0.73 & 1.18 & 1.29 & 0.68 & 0.77 & 0.90 & 0.42 & 0.66 & 0.49 & 0.74 & 30m \\

& RaDe-GS~\cite{Zhang2024ARXIV}
& 0.46 & 0.73 & \cellcolor{yellow!40}0.33 & 0.38 & \cellcolor{orange!40}0.79 & 0.75 & 0.76 & 1.19 & 1.22 & 0.62 & 0.70 & 0.78 & 0.36 & 0.68 & 0.47 & 0.68 & 8.3m \\

& PGSR~\cite{Chen2024ARXIV}
& \cellcolor{red!30}0.34
& \cellcolor{orange!40}0.58
& \cellcolor{orange!40}0.29
& \cellcolor{red!30}0.29
& \cellcolor{red!30}0.78
& \cellcolor{yellow!40}0.58
& 0.54
& \cellcolor{red!30}1.01
& \cellcolor{orange!40}0.73
& \cellcolor{red!30}0.51
& \cellcolor{yellow!40}0.49
& \cellcolor{yellow!40}0.69
& \cellcolor{red!30}0.31
& \cellcolor{orange!40}0.37
& \cellcolor{orange!40}0.38
& \cellcolor{red!30}0.53
& 60m \\

\cmidrule(lr){1-2} \cmidrule(lr){3-17} \cmidrule(lr){18-19}

% --- mesh section ---
\multirow{3}{*}{\rotatebox{90}{mesh}}

& \rev{Triangle Splatting~\cite{Held_2026_3DV}}
& 0.98 & 1.07 & 1.07 & 0.51 & 1.67 & 1.44 & 1.17 & 1.32 & 1.75 & 0.98 & 0.96 & 1.11 & 0.56 & 0.93 & 0.72 & 1.06 & --- \\

& \rev{MeshSplatting~\cite{Held_2026_CVPR}} &
 0.77 &  \cellcolor{yellow!40}0.72 & 0.74 & 0.60 &0.89 & 1.00 & 0.81 & \cellcolor{yellow!40}1.09 & 1.19 & \cellcolor{yellow!40}0.58 & 0.68 & 0.93 & 0.63 & 0.66 & 0.59 & 0.79 & --- \\

 & MILo~\cite{Guedon2025TOG}
& 0.43
& 0.74
& 0.34
& \cellcolor{yellow!40}0.37
& 0.80
& 0.74
& 0.70
& 1.21
& 1.22
& 0.66
& 0.62
& 0.80
& 0.37
& 0.76
& 0.48
& 0.68
& 25m \\

& \rev{Mesh Splatting}~\cite{zhang2026mesh}
& 0.46 & 0.73 & 0.49 & 0.43 & \cellcolor{yellow!40}0.77 & 0.82 & 0.65 & \cellcolor{orange!40}1.03 & 0.95 & \cellcolor{orange!40}0.52 & 0.58 & \cellcolor{red!30}0.59 & 0.37 & 0.44 & 0.42 & 0.62 & 23m \\

& IMLS-Splatting~\cite{Yang2025IMLSSplattingEM}
& 0.52 & 1.02 & 0.37 & \cellcolor{orange!40}0.32 & 0.86 & \cellcolor{red!30}0.50 & \cellcolor{red!30}0.48 &  1.15 & \cellcolor{yellow!40} 0.76 & 0.59 & \cellcolor{red!30}0.37 &\cellcolor{orange!40} 0.67 & \cellcolor{yellow!40}0.33 & \cellcolor{yellow!40}0.33 & \cellcolor{yellow!40}0.34 & \cellcolor{orange!40}0.57 & 11 m \\

& Ours
& \cellcolor{yellow!40}0.41
& \cellcolor{red!30}0.54
& \cellcolor{red!30}0.28
& \cellcolor{orange!40}0.32
& 0.93
& \cellcolor{yellow!40}0.58
& \cellcolor{orange!40}0.50
& 1.16
& \cellcolor{red!30}0.71
& \cellcolor{yellow!40}0.58
& \cellcolor{red!30}0.37
& \cellcolor{red!30}0.59
& \cellcolor{yellow!40}0.33
& \cellcolor{red!30}0.32
& \cellcolor{red!30}0.33
& \cellcolor{red!30}0.53
& 30m* \\

\midrule
\end{tabular}

\end{table*}

\begin{table*}[t]
\caption{\textbf{Quantitative comparison of geometric reconstruction on the BlendedMVS dataset.} We report Chamfer distance (mm, $\downarrow$) for 18 test scenes. Colors indicate \colorbox{red!30}{1st}, \colorbox{orange!40}{2nd}, and \colorbox{yellow!40}{3rd} place per scene. Our method achieves the best mean Chamfer distance among mesh-based methods.}
\label{tab:bmvs_results}
\centering
\resizebox{1.0\linewidth}{!}{
\begin{tabular}{l|cccccccccccccccccc|c}
\toprule
Method & Basketball & Bear & Bread & Camera & Clock & Cow & Dog & Doll & Dragon & Durian & Fountain & Gundam & House & Jade & Man & Monster & Sculpture & Stone & Mean \\
\midrule
NeuS           & 2.96 & 3.00 & 2.85 & 2.61 & 2.75 & 2.04 & 2.75 & 2.17 & 2.95 & 3.14 & 3.03 & 1.62 & 3.23 & 4.25 & 2.29 & 1.92 & 2.10 & 2.51 & 2.68 \\

Surfels        & 1.59 & 1.52 & 1.41 & 1.75 & 5.08 & 3.21 & 2.97 & 2.32 & 2.93 & 4.01 & 2.65 & 0.97 & 1.76 & 3.49 & 2.23 & 1.36 & 3.07 & 2.02 & 2.46 \\

SuGar          & 8.00 & 9.73 & 7.65 & 7.77 & 9.21 & 8.69 & 9.27 & 8.75 & 9.76 & 8.04 & 9.05 & 7.32 & 7.28 & 10.7 & 9.29 & 8.20 & 8.98 & 9.19 & 8.71 \\

PGSR           & \cellcolor{yellow!40}0.92 & \cellcolor{orange!40}1.05 & \cellcolor{red!30}0.74 & \cellcolor{yellow!40}1.51 & \cellcolor{red!30}1.64 & \cellcolor{red!30}1.09 & \cellcolor{red!30}1.35 & \cellcolor{red!30}1.33 & \cellcolor{yellow!40}1.51 & \cellcolor{orange!40}1.63 & \cellcolor{red!30}1.50 & 0.92 & \cellcolor{red!30}1.18 & \cellcolor{orange!40}3.01 & \cellcolor{red!30}1.17 & \cellcolor{orange!40}0.97 & \cellcolor{red!30}1.25 & \cellcolor{red!30}1.08 & \cellcolor{red!30}1.35 \\

\midrule

IMLS-Splatting & 2.48 & 1.86 & 2.69 & 3.61 & 2.96 & 2.80 & 2.85 & 2.32 & 2.39 & 3.35 & 2.83 & 1.78 & 3.01 & 5.10 & 2.61 & 1.99 & 2.04 & 2.85 & 2.75 \\

\rev{Mesh Splatting}  & 1.26 & 1.16 & \cellcolor{orange!40}0.75 & 1.75 & \cellcolor{orange!40}1.99 & \cellcolor{yellow!40}1.30 & \cellcolor{yellow!40}2.08 & 2.02 & 1.57 &  2.36 & \cellcolor{yellow!40}2.37 & \cellcolor{yellow!40}0.90 & 1.94 & 3.27 & \cellcolor{yellow!40}1.52 & 1.30 & \cellcolor{yellow!40}1.64 & 1.64 & 1.71 \\

MILo           & \cellcolor{red!30}0.83 & \cellcolor{red!30}0.95 & 1.46 & \cellcolor{red!30}1.01 & \cellcolor{yellow!40}2.03 & 1.38 & 2.23 & \cellcolor{orange!40}1.42 & \cellcolor{red!30}1.36 & \cellcolor{red!30}1.35 & 2.47 & \cellcolor{orange!40}0.79 & \cellcolor{orange!40}1.45 & \cellcolor{yellow!40}3.02 & 2.90 & \cellcolor{yellow!40}1.13 & 2.31 & \cellcolor{orange!40}1.09 & \cellcolor{yellow!40}1.62 \\

Ours           & \cellcolor{orange!40}0.91 & \cellcolor{yellow!40}1.14 & \cellcolor{yellow!40}0.95 & \cellcolor{orange!40}1.19 & 2.60 & \cellcolor{orange!40}1.23 & \cellcolor{orange!40}1.73 & \cellcolor{yellow!40}1.44 & \cellcolor{orange!40}1.38 & \cellcolor{yellow!40}2.19 & \cellcolor{orange!40}1.71 & \cellcolor{red!30}0.68 & \cellcolor{yellow!40}1.70 & \cellcolor{red!30}2.85 & \cellcolor{orange!40}1.35 & \cellcolor{red!30}0.92 & \cellcolor{orange!40}1.27 & \cellcolor{yellow!40}1.28 & \cellcolor{orange!40}1.47 \\

% Ours (mono)   & 0.84 & 1.02 & 0.80 & 1.20 & 2.48 & 1.25 & 1.54 & 1.41 & 1.36 & 1.93 & 1.70 & 0.69 & 1.40 & 2.73 & 1.34 & 0.77 & 1.29 & 1.38 & 1.40 \\

\bottomrule
\end{tabular}
}

\end{table*}

\subsection{Datasets and Metrics}
We evaluate our method on common reconstruction datasets: \textbf{NeRF Synthetic}~\cite{Mildenhall2020ECCV}, \textbf{DTU}~\cite{Jensen2014CVPR}, and \textbf{BlendedMVS}~\cite{yao2020blendedmvs}. The NeRF Synthetic dataset consists of eight synthetic objects rendered in Blender. The DTU dataset comprises real-world object scans captured in a controlled indoor environment with fixed lighting and calibrated camera poses. The BlendedMVS dataset consists of in-the-wild captures with intricate shapes. In the supplementary material, we additionally report on the \textbf{Stanford-ORB}~\cite{Kuang2023NEURIPS} benchmark from which we select the five most complex objects.
Following existing work, we report the \textbf{Chamfer distance} for 15 scenes of the DTU dataset and 18 scenes from the low-res set of the BlendedMVS dataset, using official evaluation protocols. For the NeRF Synthetic dataset, we report the PSNR and SSIM metrics. 

\subsection{Results}
\begin{figure*}[h]
  \centering
  \includegraphics[width=1.0\linewidth]{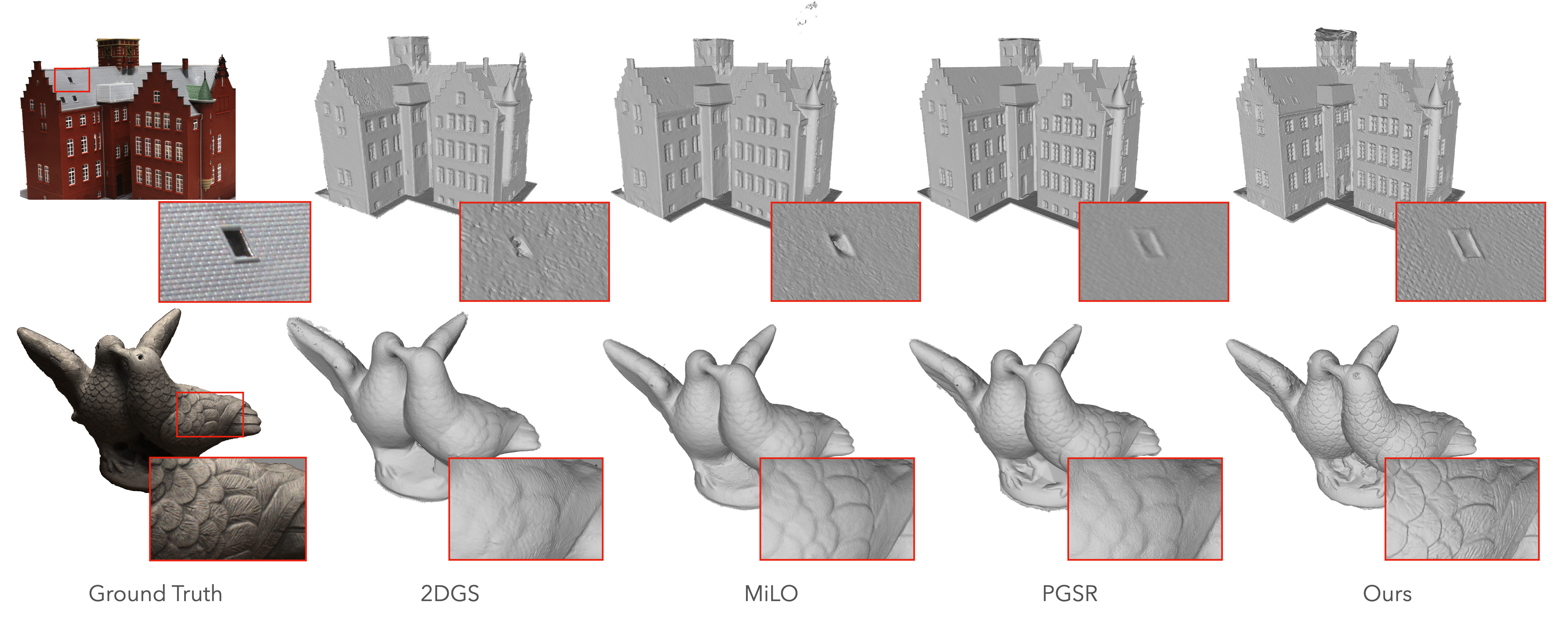}
    \caption{\textbf{Qualitative comparison on the DTU dataset.} We compare mesh reconstructions on two challenging scenes. 2DGS and MiLO produce oversmoothed surfaces that lose geometric detail. PGSR captures some details but exhibits surface noise. Our method reconstructs sharp edges and high-frequency details while maintaining clean mesh quality, as visible in the roof tiles and the bird's textured feathers.}
   \label{fig:dtu_results}
\end{figure*}

\paragraph{DTU Dataset} Table~\ref{tab:dtu_results} reports the Chamfer distances of meshes reconstructed by our method compared to a range of baselines based on implicit representations, Gaussian splatting, and native mesh representations across 15 scenes from the DTU dataset. 
We caution that the DTU ground truth point clouds are incomplete, failing to capture the backsides of target objects due to the limited coverage of captured views. This missing geometry naturally biases the evaluation metric in favor of methods that produce incomplete or open meshes, such as TSDF fusion-based approaches like PGSR. To ensure a fair comparison, we apply a straightforward filtering procedure to our results, removing any triangles from our reconstructed meshes that are not visible in at least two input views. We provide a more detailed discussion regarding this dataset bias in the supplementary material. 
Quantitatively, our approach significantly outperforms the state-of-the-art mesh-based method, MILo, and performs on par with the overall state-of-the-art approach, PGSR. Qualitatively, our mesh-based reconstruction excels at capturing high-curvature geometry and sharp edges while competing methods tend to produce oversmooth meshes with missing details (Fig.~\ref{fig:dtu_results}). 

\begin{figure*}[p]
  \centering
  \includegraphics[width=1.0\linewidth]{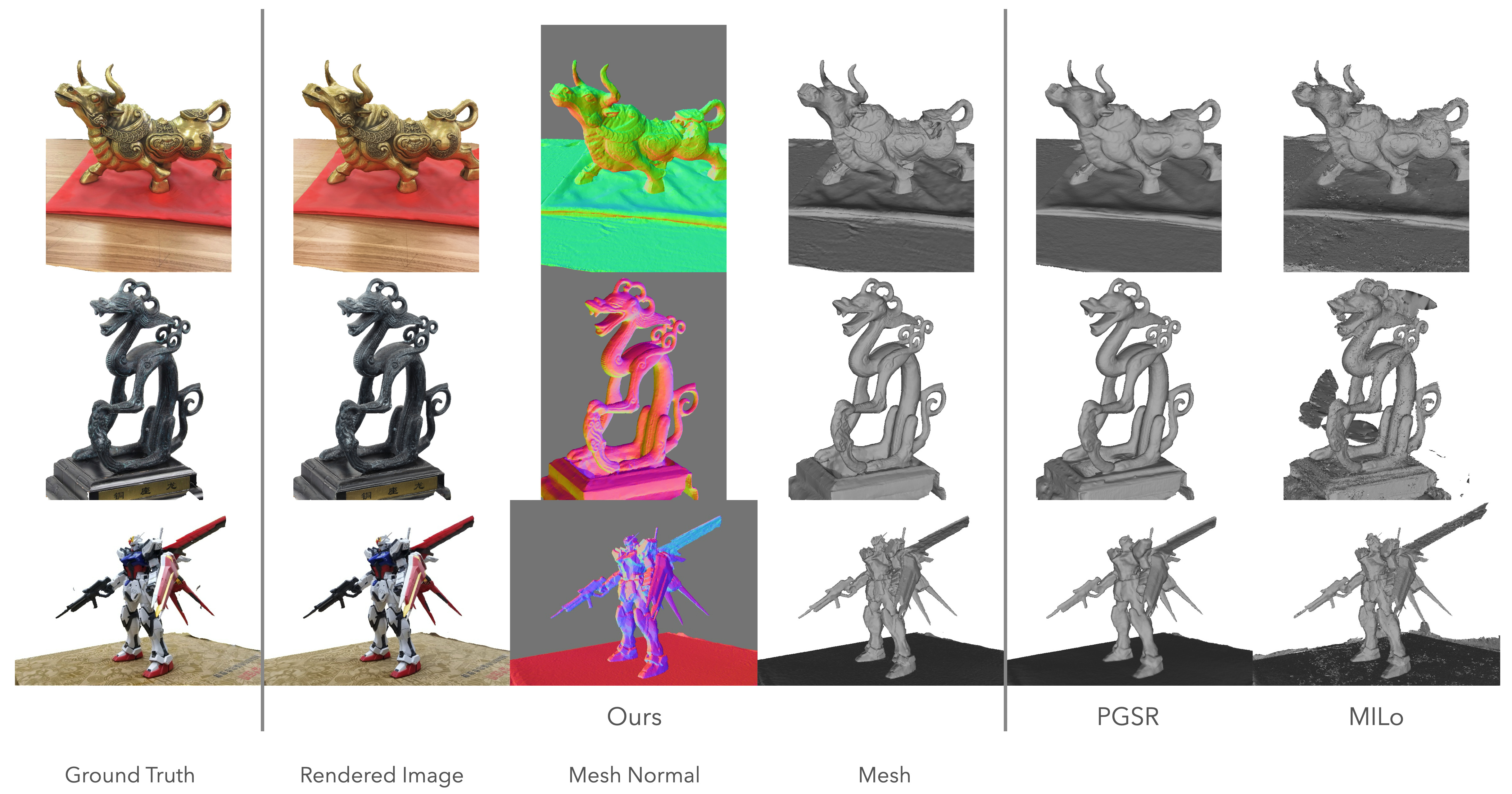}
    \caption{\textbf{Qualitative comparison on the BlendedMVS dataset.} For each scene, we show the ground truth image, our rendered image, our predicted mesh normals, and mesh reconstructions from our method, PGSR, and MiLO. Our method produces detailed meshes that faithfully capture object silhouettes and detailed surface normals. PGSR produces overly smooth surfaces, while MILo suffers from high frequency artefacts. Best viewed zoomed in.}
   \label{fig:bmvs_results}
\end{figure*}

\paragraph{BlendedMVS} Table~\ref{tab:bmvs_results} reports the Chamfer distance of mesh reconstruction methods across 18 scenes from the BlendedMVS dataset. 
We observed that both MILo and PGSR diverge on several instances of the BlendedMVS and Stanford ORB datasets when trained without foreground masks. To ensure a fair comparison, we augmented both baselines with an auxiliary $L_2$ mask loss term identical to the one employed in our optimization. We set the loss weight to $\lambda_{mask}=0.1$ for both competing methods, maintaining a consistent ratio between the photometric image loss and the mask loss when compared to our method. Our method yields significant improvements over the closest mesh-based reconstruction methods. These quantitative gains are further supported by the qualitative comparisons shown in Fig.~\ref{fig:bmvs_results}. \rev{Our method requires roughly 32 GB of peak GPU memory, making it more memory-intensive than competing approaches. Reducing the batch size lowers the memory footprint below 24 GB and cuts runtime to 14 minutes with only a marginal loss in quality. A more detailed discussion of this trade-off is provided in the supplementary material.}

\begin{figure*}[h]
  \centering
  \includegraphics[trim={0pt 80pt 0pt 10pt}, clip, width=1.0\linewidth]{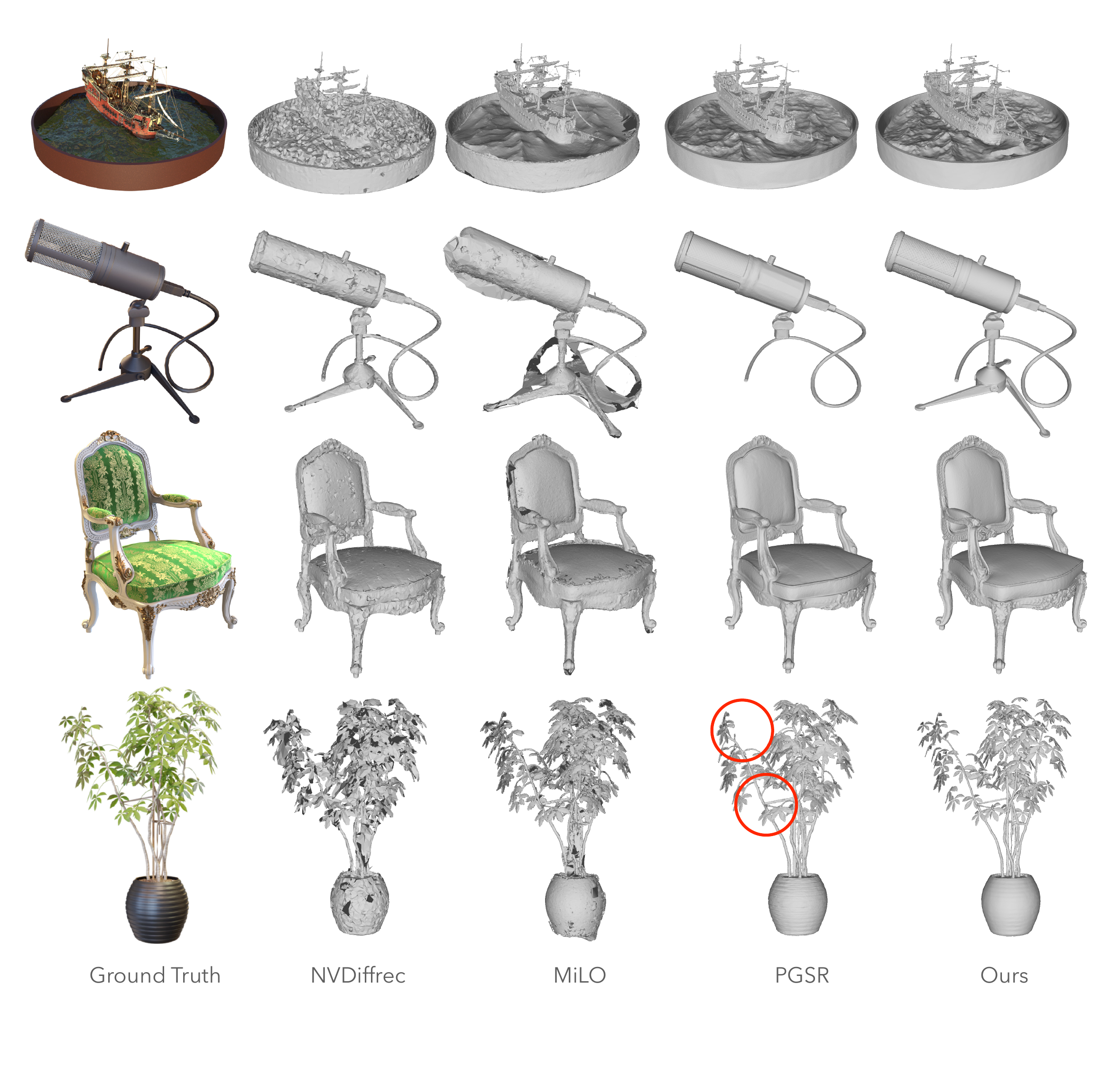}
    \caption{\textbf{Qualitative comparison on the NeRF Synthetic dataset.} We compare mesh reconstructions from NVDiffrec, MILO, PGSR, and our method. NVDiffrec produces noisy surfaces with limited geometric detail. MILO reconstructs more detailed meshes but exhibit artifacts around the object silhouette. PGSR is lacking fine-grained details and additionally suffers from missing geometry due to its post processing steps. Our method recovers fine-grained geometry, such as individual leaves and the microphone grille, while maintaining clean surface quality.}
   \label{fig:nerf_results}
\end{figure*}

\paragraph{NeRF Synthetic}
We also evaluate the performance our method on the NeRF Synthetic dataset. Quantitative results are reported in Table~\ref{tab:nerf_results}. For each method we report two sets of metrics: (i) the method's native renderer (3DGS-based for PGSR and MILo, mesh rasterization with neural shading for ours), and (ii) novel-view synthesis from the extracted mesh shaded only with the per-vertex diffuse albedo. For our method, the per-vertex diffuse color is obtained by uniformly sampling 32 hemisphere directions around the surface normal, querying the appearance model along each outgoing direction, and taking the cosine-weighted average. 
%The drop in PSNR between native and mesh-only rendering is markedly smaller for our method than for the baselines, and we outperform both PGSR and MILO when evaluating the rendered extracted meshes.
The drop in PSNR between native rendering and mesh-only rendering is substantially smaller for ADELE than for PGSR and MILo. This suggests that ADELE captures scene fidelity directly in the reconstructed geometry, whereas competing approaches rely more heavily on learned rendering representations to reproduce fine-scale appearance details. Qualitative comparisons (Fig.~\ref{fig:nerf_results}) further confirm that our method recovers fine-grained geometric detail more faithfully than the competing approaches. Additional novel-view synthesis results are provided in the supplementary material.

\begin{table}[h]
\caption{\textbf{Quantitative comparison on NeRF Synthetic.} PSNR ($\uparrow$) and SSIM ($\uparrow$) under two rendering protocols: \emph{Native rendering} uses each method's own renderer (3DGS for PGSR/MiLO, neural-shaded mesh rasterization for ours); \emph{Extracted-mesh rendering} renders the extracted mesh shaded with a per-vertex diffuse albedo.}
\label{tab:nerf_results}
\centering
\resizebox{\linewidth}{!}{
\begin{tabular}{ll|ccccccccc}
\multicolumn{2}{c}{} & \multicolumn{9}{c}{PSNR$\uparrow$} \\
Render & Method & Chair & Drums & Ficus & Hotdog & Lego & Mats. & Mic & Ship & Avg \\
\hline
\multirow{3}{*}{Native}
       & PGSR & 31.62 & 24.69 & 28.60 & 35.09 & 31.67 & 28.93 & 33.81 & 29.14 & 30.44 \\
       & MILo & 32.69 & 25.52 & 34.67 & 34.82 & 33.77 & 29.11 & 33.51 & 30.76 & 31.85 \\
      & NVDiffrec & 31.60 & 24.10 & 30.88 & 33.04 & 29.14 & 26.74 & 30.78 & 26.12 & 29.05 \\
       & Ours & 31.68 & 24.42 & 29.48 & 34.00 & 30.79 & 26.41 & 32.25 & 28.43 & 29.68 \\
\hline
\multirow{4}{*}{Mesh}
       & PGSR             & 27.12 & 16.40 & 22.13 & 29.60 & 27.41 & 14.86 & 24.95 & \textbf{24.69} & 23.40 \\
       & MILo             & 24.28 & 21.74 & 24.39 & 28.56 & 24.98 & 22.00 & 24.80 & 24.67 & 24.43 \\
       & Ours             & \textbf{28.61} & \textbf{23.14} & \textbf{27.05} & \textbf{30.64} & \textbf{29.70} & \textbf{22.20} & \textbf{26.91} & 23.49 & \textbf{26.47} \\
\multicolumn{11}{c}{} \\[-2mm]
\multicolumn{2}{c}{} & \multicolumn{9}{c}{SSIM$\uparrow$} \\
Render & Method & Chair & Drums & Ficus & Hotdog & Lego & Mats. & Mic & Ship & Avg \\
\hline
\multirow{3}{*}{Native}
       & PGSR & 0.981 & 0.939 & 0.948 & 0.981 & 0.971 & 0.943 & 0.985 & 0.900 & 0.956 \\
       & MILo & 0.980 & 0.951 & 0.986 & 0.981 & 0.977 & 0.955 & 0.987 & 0.902 & 0.965 \\
      & NVDiffrec & 0.969 & 0.916 & 0.970 & 0.973 & 0.949 & 0.923 & 0.977 & 0.833 & 0.939 \\
       & Ours & 0.978 & 0.939 & 0.957 & 0.976 & 0.964 & 0.915 & 0.979 & 0.867 & 0.946 \\
\hline
\multirow{4}{*}{Mesh}
       & PGSR             & 0.939 & 0.848 & 0.899 & 0.958 & 0.921 & 0.723 & 0.929 & \textbf{0.812} & 0.879 \\
       & MILo             & 0.915 & 0.874 & 0.901 & 0.947 & 0.891 & 0.858 & 0.898 & 0.795 & 0.885 \\
       & Ours             & \textbf{0.958} & \textbf{0.916} & \textbf{0.946} & \textbf{0.964} & \textbf{0.953} & \textbf{0.865} & \textbf{0.948} & 0.801 & \textbf{0.919} \\
\hline
\end{tabular}
}
\end{table}

\begin{table}[h]
\centering
\caption{\textbf{Ablation studies on the DTU and BMVS datasets.}
We report the Chamfer Distance (↓) for different variants of our method on both datasets.}
\label{tab:ablation_results}

\setlength{\tabcolsep}{3pt}
\renewcommand{\arraystretch}{0.95}
\begin{tabular}{@{}lcc@{}}
\toprule
\textbf{Configuration} & \textbf{DTU}\,↓ & \textbf{BMVS}\,↓ \\
\midrule
Full Model (ours)        & 0.53 & 1.47 \\
w/o Mesh-Based Rendering & 0.79 & 3.17 \\
w/o Hybrid Training      & 0.72 & 1.98 \\
w/o Densification        & 0.59  & 5.37\\
w/o Offset Sampling      & 0.55 & 1.56 \\
\bottomrule
\end{tabular}
\end{table}

\subsection{Ablation Studies}
\label{subsec:ablation_studies}

% \begin{table}[t]
% \centering
% \caption{\textbf{Ablation studies on the DTU and BMVS datasets.} 
% We report the Chamfer Distance (↓) for different variants of our method on both datasets. Lower is better. Each component contributes to improving the geometric accuracy of the reconstructed meshes.}
% \label{tab:ablation_results}
% \begin{tabular}{l cc}
% \toprule
% \textbf{Configuration} & \textbf{DTU} ↓ & \textbf{BMVS} ↓ \\
% \midrule
% Full Model (ours)              & 0.53 & 1.47 \\
% w/o Mesh-Based Rendering       & xx & xx \\
% w/o Hybrid Training       &  0.72 & 1.98 \\
% w/o Densification              & xx & xx \\
% w/o Offset Sampling            & 0.55 & 1.56 \\
% %w/o Adaptive Grid Positions    & 0.55 & xx \\
% %w/o Grid Updates               & 0.88 & xx \\
% %w/o Hybrid Training (VGGT init.)       &  & xx \\
% \bottomrule
% \end{tabular}
% \end{table}

\begin{figure*}[p]
  \centering
  \includegraphics[width=1.0\linewidth]{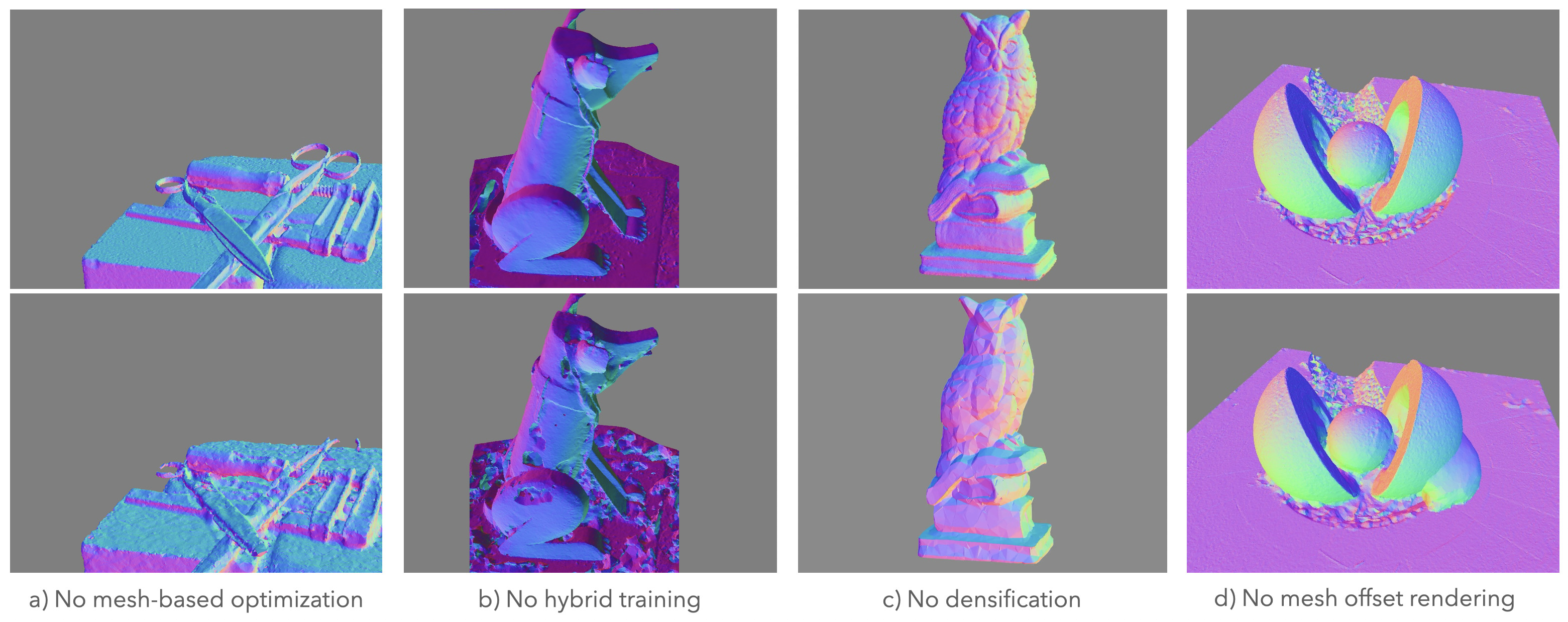}
    \caption{Ablational results on scenes of the DTU and BlendedMVS datasets. The top row shows results produced by our full method, while the bottom row presents results from different ablation variants.}
   \label{fig:ablation_results}
\end{figure*}

We conduct extensive ablation studies on the DTU and BMVS datasets to evaluate the contribution of individual components in our method, the results of which are summarized in Table~\ref{tab:ablation_results} and Figure~\ref{fig:ablation_results}.

\noindent\textbf{Mesh-Based Rendering.}
We evaluate the benefit of mesh-based rendering within our optimization pipeline by training an SDF hash-grid using only volumetric rendering following the NeuS formulation, accelerated with a learned occupancy grid. We then extract meshes using our adaptively triangulated grid, initialized with 250k uniformly sampled points and identical densification steps to our optimization routine. The resulting reconstructions show incomplete and messy geometry with missing parts of the scene.

\noindent\textbf{Hybrid Training.}
To ablate the proposed hybrid training strategy, 
%which combines volumetric and mesh-based rendering supervision, 
we evaluate a variant without volumetric training. Several scenes exhibit severe artifacts, particularly under spherical initialization. This highlights the sensitivity of purely mesh-based optimization to initialization quality and demonstrates the importance of the volumetric rendering signal for stabilizing the optimization process.

\noindent\textbf{Grid Densification and Pruning.}
We analyze the effect of our locally refinable tetrahedral grid obtained via Delaunay triangulation by evaluating a variant of the method in which only the initial grid vertices are optimized, without applying the proposed gradient-guided refinement strategy. The visual comparison clearly shows that the extracted meshes are overly simplified and lack detail.

\noindent\textbf{Offset Sampling.}
\rev{While offset sampling incurs a modest increase in computational cost due to the additional samples required per batch, it reduces scene-specific failure cases, most prominently convex artifacts. Its contribution to the aggregate metrics is small, but it leads to noticeable qualitative improvements in the affected scenes. Additional examples are provided in the supplementary material.}

\subsection{Large Scale Scenes}
\label{subsec:large_scale_scenes}

\rev{For unbounded scenes, the spherical initialization used in the object-centric case is insufficient. We instead initialize from a coarse SDF derived from the predicted depth of a feed-forward model (VGGT) and learn a residual field relative to this base geometry. Details of the initialization and hyperparameters are included in the supplementary material.  \autoref{fig:tnt_results} shows qualitative results on the Tanks and Temples dataset. Our method recovers high-fidelity geometry in the primary subject areas, but remains confined to a bounded region of interest at the scene center, leaving background areas unmodeled.
}

\begin{figure}[h]
  \centering
   \includegraphics[width=1.0\linewidth]{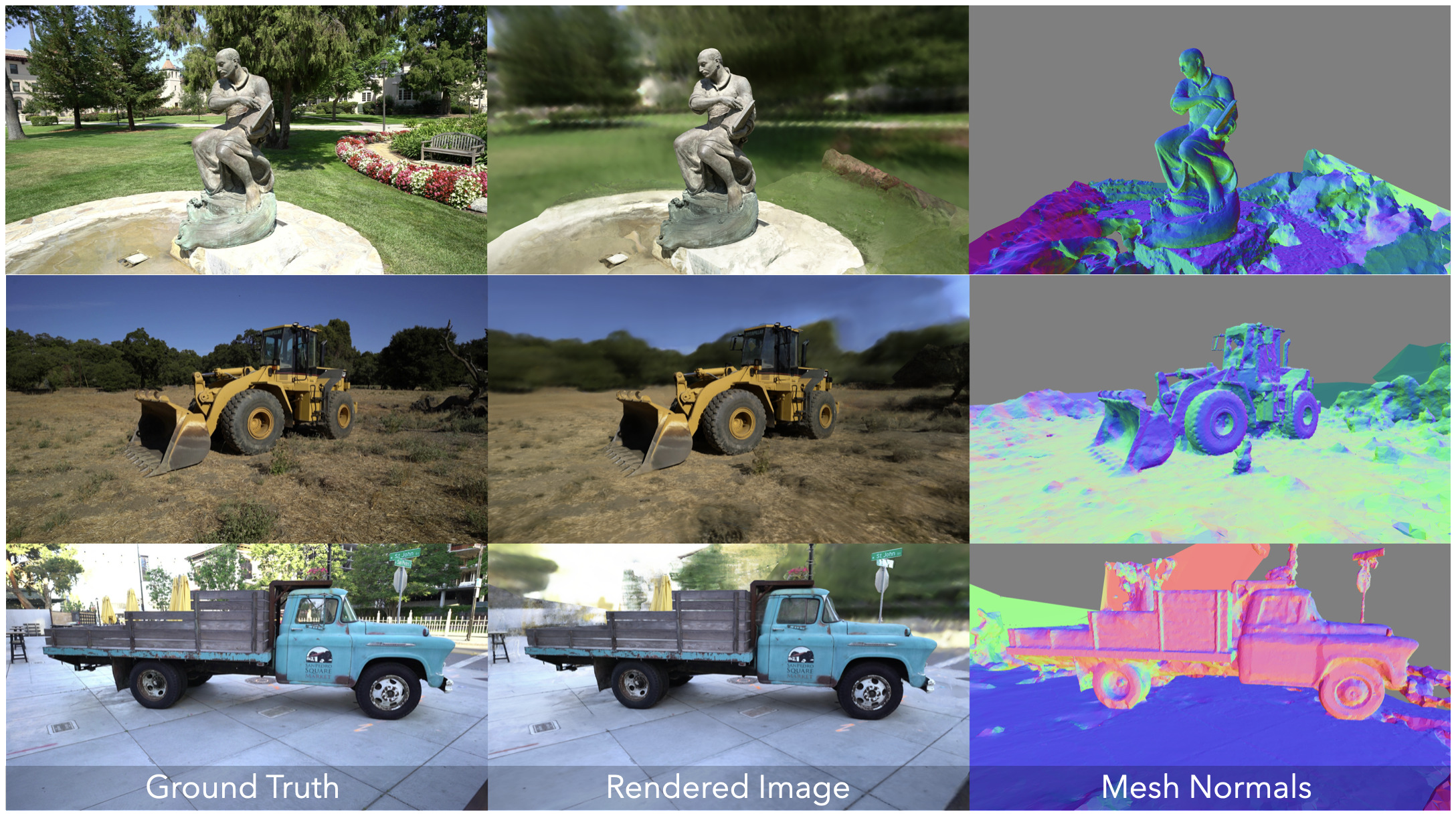}
   \caption{\rev{Examples of mesh reconstructions obtained with our method on the TnT dataset.}}
   \label{fig:tnt_results}
\end{figure}

% \noindent\textbf{Adaptive Grid Positions.}
% We analyze the effect of our adaptive tetrahedral grid obtained via Delaunay triangulation by comparing it against a variant in which the grid points remain fixed and only the SDF is optimized. While this variant performs as well as our base method in the metric evaluation, we observe that constraining the grid points limits the representational capacity of the model, leading to a qualitative degradation in surface smoothness.

\section{Conclusion}
\label{sec:conclusion}
Our evaluation across various object-centric benchmarks demonstrates that direct mesh optimization, without reliance on auxiliary explicit primitives, serves as a viable alternative for 3D reconstruction. Notably, our method excels in recovering fine-grained details and sharp edges, highlighting the inherent difficulties that diffuse or density-based representations face when reconstructing such high-frequency features. \rev{However, the current pipeline still faces some limitations. It requires more memory than competing baselines, remains susceptible to geometric artifacts in regions with high specularity, and is restricted to a region of interest in unbounded scenes. We believe that incorporating 
%more sophisticated differentiable rendering techniques beyond our current sampling-based scheme or 
stronger initialization techniques and improved appearance models could effectively mitigate these ambiguities.}

% However, we acknowledge that the current pipeline remains susceptible to geometric artifacts in regions with complex appearance or high specularity. We believe that incorporating more sophisticated differentiable rendering techniques beyond our current sampling-based scheme or integrating improved appearance models could effectively mitigate these ambiguities.

\clearpage
\bibliographystyle{ACM-Reference-Format}
\bibliography{main}

@String(CVPR= {IEEE Conf. Comput. Vis. Pattern Recog.})

@String(ICCV= {Int. Conf. Comput. Vis.})

@String(ECCV= {Eur. Conf. Comput. Vis.})

@String(TOG= {ACM Trans. Graph.})

@String(ICLR = {Int. Conf. Learn. Represent.})

@String(CVPR  = {CVPR})

@String(ICCV  = {ICCV})

@String(ECCV  = {ECCV})

@String(TOG   = {ACM TOG})

@String(ICLR  = {ICLR})

@String(NEURIPS = {NeurIPS})

@STRING{SIGGRAPH = {ACM Trans. on Graphics}}

@inproceedings{Mildenhall2020ECCV,
  author    = {Mildenhall, Ben and Srinivasan, Pratul P and Tancik, Matthew and Barron, Jonathan T and Ramamoorthi, Ravi and Ng, Ren},
  title     = {{NeRF}: Representing scenes as neural radiance fields for view synthesis},
  booktitle = ECCV,
  year      = {2020}
}

@article{Barron2022CVPR,
  title     = {Mip-NeRF 360: Unbounded Anti-Aliased Neural Radiance Fields},
  author    = {Jonathan T. Barron and Ben Mildenhall and Dor Verbin and Pratul P. Srinivasan and Peter Hedman},
  journal   = {CVPR},
  year      = {2022}
}

@inproceedings{Yu2022CVPR,
  author    = {{Alex Yu and Sara Fridovich-Keil} and Matthew Tancik and Qinhong Chen and Benjamin Recht and Angjoo Kanazawa},
  title     = {Plenoxels: Radiance Fields without Neural Networks},
  booktitle = CVPR,
  year      = {2022}
}

@article{Mueller2022TOG,
  author    = {Thomas M\"uller and Alex Evans and Christoph Schied and Alexander Keller},
  title     = {Instant Neural Graphics Primitives with a Multiresolution Hash Encoding},
  journal   = {ACM transactions on graphics (TOG)},
  volume    = {41},
  number    = {4},
  pages     = {1--15},
  year      = {2022},
  publisher = {ACM New York, NY, USA}
}

@inproceedings{Chen2022ECCV,
  author    = {Anpei Chen and Zexiang Xu and Andreas Geiger and Jingyi Yu and Hao Su},
  title     = {TensoRF: Tensorial Radiance Fields},
  booktitle = ECCV,
  year      = {2022}
}

@inproceedings{Kulhanek2023ICCV,
  title     ={{T}etra-{NeRF}: Representing Neural Radiance Fields Using Tetrahedra},
  author    = {Kulhanek, Jonas and Sattler, Torsten},
  booktitle = ICCV,
  year      = {2023}
}

@inproceedings{Govindarajan2025ICCV,
  author    = {Govindarajan, Shrisudhan and Rebain, Daniel and Yi, Kwang Moo and Tagliasacchi, Andrea},
  title     = {Radiant Foam: Real-Time Differentiable Ray Tracing},
  booktitle = ICCV,
  year      = {2025}
}

@article{Kerbl2023TOG,
  title     = {3d gaussian splatting for real-time radiance field rendering},
  author    = {Kerbl, Bernhard and Kopanas, Georgios and Leimk{\"u}hler, Thomas and Drettakis, George},
  journal   = {ACM Trans. Graph.},
  volume    = {42},
  number    = {4},
  pages     = {139--1},
  year      = {2023}
}

@inproceedings{Barron2021ICCV,
  author    = {Barron, Jonathan T. and Mildenhall, Ben and Tancik, Matthew and Hedman, Peter and Martin-Brualla, Ricardo and Srinivasan, Pratul P.},
  title     = {Mip-NeRF: A Multiscale Representation for Anti-Aliasing Neural Radiance Fields},
  booktitle = ICCV,
  year      = {2021}
}

@inproceedings{MartinBrualla2021CVPR,
  author    = {Martin-Brualla, Ricardo and Radwan, Noha and Sajjadi, Mehdi S. M. and Barron, Jonathan T. and Dosovitskiy, Alexey and Duckworth, Daniel},
  title     = {NeRF in the Wild: Neural Radiance Fields for Unconstrained Photo Collections},
  booktitle = CVPR,
  year      = {2021}
}

@inproceedings{Sun2022CVPR,
  author    = {Sun, Cheng and Sun, Min and Chen, Hwann-Tzong},
  title     = {Direct Voxel Grid Optimization: Super-Fast Convergence for Radiance Fields Reconstruction},
  booktitle = CVPR,
  year      = {2022}
}

@inproceedings{Hu2023ICCV,
  author    = {Hu, Wenbo and Wang, Yuling and Ma, Lin and Yang, Bangbang and Gao, Lin and Liu, Xiao and Ma, Yuewen},
  title     = {Tri-MipRF: Tri-Mip Representation for Efficient Anti-Aliasing Neural Radiance Fields},
  booktitle = ICCV,
  year      = {2023}
}

@inproceedings{Fang2024ECCV,
  author    = {Fang, Guangchi and Wang, Bing},
  title     = {Mini-Splatting: Representing Scenes with a Constrained Number of Gaussians},
  booktitle = ECCV,
  year      = {2024}
}

@inproceedings{Yu2024CVPR,
  author    = {Yu, Zehao and Chen, Anpei and Huang, Binbin and Sattler, Torsten and Geiger, Andreas},
  title     = {Mip-Splatting: Alias-free 3D Gaussian Splatting},
  booktitle = CVPR,
  year      = {2024}
}

@inproceedings{Liang2024ECCV,
  author    = {Liang, Zhihao and Zhang, Qi and Hu, Wenbo and Feng, Ying and Zhu, Lei and Jia, Kui},
  title     = {Analytic-Splatting: Anti-Aliased 3D Gaussian Splatting via Analytic Integration},
  booktitle = ECCV,
  year      = {2024}
}

@article{Fang2024ARXIV,
  author  = {Fang, Guangchi and Wang, Bing},
  title   = {Mini-Splatting2: Building 360 Scenes within Minutes via Aggressive Gaussian Densification},
  journal = {arXiv.org},
  volume  = {2411.12788},
  year    = {2024}
}

@article{Ye2024JMLR,
  author  = {Ye, Vickie and Li, Ruilong and Kerr, Justin and Turkulainen, Matias and Yi, Brent and Pan, Zhuoyang and Seiskari, Otto and Ye, Jianbo and Hu, Jeffrey and Tancik, Matthew and Kanazawa, Angjoo},
  title   = {gsplat: An Open-Source Library for Gaussian Splatting},
  journal = {Journal of Machine Learning Research},
  volume  = {26},
  year    = {2024}
}

@inproceedings{Kheradmand2025ARXIV,
  title={StochasticSplats: Stochastic Rasterization for Sorting-Free 3D Gaussian Splatting},
  author={Kheradmand, Shakiba and Vicini, Delio and Kopanas, George and Lagun, Dmitry and Yi, Kwang Moo and Matthews, Mark and Tagliasacchi, Andrea},
  booktitle={Proceedings of the IEEE/CVF International Conference on Computer Vision (ICCV)},
  year={2025}
}

@inproceedings{Oechsle2021ICCV,
  author    = {Michael Oechsle and Songyou Peng and Andreas Geiger},
  title     = {UNISURF: Unifying Neural Implicit Surfaces and Radiance Fields for Multi-View Reconstruction},
  booktitle = ICCV,
  year      = {2021}
}

@inproceedings{Wang2021NEURIPS,
  author    = {Peng Wang and Lingjie Liu and Yuan Liu and Christian Theobalt and Taku Komura and Wenping Wang},
  title     = {NeuS: Learning Neural Implicit Surfaces by Volume Rendering for Multi-view Reconstruction},
  booktitle = NEURIPS,
  year      = {2021}
}

@inproceedings{Yariv2021NEURIPS,
  author    = {Yariv, Lior and Gu, Jiatao and Kasten, Yoni and Lipman, Yaron},
  title     = {Volume rendering of neural implicit surfaces},
  booktitle = NEURIPS,
  year      = {2021}
}

@inproceedings{Wang2023ICCV,
  title     = {NeuS2: Fast Learning of Neural Implicit Surfaces for Multi-view Reconstruction},
  author    = {Wang, Yiming and Han, Qin and Habermann, Marc and Daniilidis, Kostas and Theobalt, Christian and Liu, Lingjie},
  booktitle = ICCV,
  year      = {2023}
}

@inproceedings{Alexandru2023CVPR,
  Title     = {PermutoSDF: Fast Multi-View Reconstruction with Implicit Surfaces using Permutohedral Lattices},
  Author    = {Radu Alexandru Rosu and Sven Behnke},
  Booktitle = CVPR,
  Year      = {2023}
}

@inproceedings{Li2023CVPR,
  title     = {Neuralangelo: High-Fidelity Neural Surface Reconstruction},
  author    = {Li, Zhaoshuo and M\"uller, Thomas and Evans, Alex and Taylor, Russell H and Unberath, Mathias and Liu, Ming-Yu and Lin, Chen-Hsuan},
  booktitle = CVPR,
  year      = {2023}
}

@inproceedings{Guedon2024CVPR,
  title     = {SuGaR: Surface-Aligned Gaussian Splatting for Efficient 3D Mesh Reconstruction and High-Quality Mesh Rendering},
  author    = {Gu{\'e}don, Antoine and Lepetit, Vincent},
  booktitle = CVPR,
  year      = {2024}
}

@article{Chen2023ARXIVa,
  title   = {NeuSG: Neural Implicit Surface Reconstruction with 3D Gaussian Splatting Guidance},
  author  = {Hanlin Chen and Chen Li and Yunsong Wang and Gim Hee Lee},
  journal = {arXiv.org},
  volume  = {2312.00846},
  year    = {2023}
}

@inproceedings{Yu2024NEURIPS,
  title     = {GSDF: 3DGS Meets SDF for Improved Rendering and Reconstruction},
  author    = {Mulin Yu and Tao Lu and Linning Xu and Lihan Jiang and Yuanbo Xiangli and Bo Dai},
  booktitle = NEURIPS,
  year      = {2024}
}

@inproceedings{Huang2024SIGGRAPH,
  title     = {2D Gaussian Splatting for Geometrically Accurate Radiance Fields},
  author    = {Huang, Binbin and Yu, Zehao and Chen, Anpei and Geiger, Andreas and Gao, Shenghua},
  publisher = {Association for Computing Machinery},
  booktitle = {SIGGRAPH 2024 Conference Papers},
  year      = {2024}
}

@inproceedings{Dai2024SIGGRAPH,
  author    = {Dai, Pinxuan and Xu, Jiamin and Xie, Wenxiang and Liu, Xinguo and Wang, Huamin and Xu, Weiwei},
  title     = {High-quality Surface Reconstruction using Gaussian Surfels},
  booktitle = {SIGGRAPH 2024 Conference Papers},
  year      = {2024}
}

@article{Yu2024TOG,
  author  = {Yu, Zehao and Sattler, Torsten and Geiger, Andreas},
  title   = {Gaussian Opacity Fields: Efficient Adaptive Surface Reconstruction in Unbounded Scenes},
  year    = {2024},
  volume  = {43},
  number  = {6},
  journal = SIGGRAPH,
  pages   = {271:1--271:13}
}

@inproceedings{Gu2024NEURIPS,
  title     = {Tetrahedron Splatting for 3D Generation},
  author    = {Gu, Chun and Yang, Zeyu and Pan, Zijie and Zhu, Xiatian and Zhang, Li},
  booktitle = NEURIPS,
  year      = {2024}
}

@inproceedings{Guedon2025CVPR,
  author    = {Gu{\'e}don, Antoine and Ichikawa, Tomoki and Yamashita, Kohei and Nishino, Ko},
  title     = {MAtCha Gaussians: Atlas of Charts for High-Quality Geometry and Photorealism From Sparse Views},
  booktitle = CVPR,
  year      = {2025}
}

@inproceedings{Wang2025CVPR,
  author    = {Wang, Jianyuan and Chen, Minghao and Karaev, Nikita and Vedaldi, Andrea and Rupprecht, Christian and Novotny, David},
  title     = {VGGT: Visual Geometry Grounded Transformer},
  booktitle = CVPR,
  year      = {2025}
}

@article{Guedon2025TOG,
  author  = {Gu{\'e}don, Antoine and Gomez, Diego and Maruani, Nissim and Gong, Bingchen and Drettakis, George and Ovsjanikov, Maks},
  title   = {MILo: Mesh-In-the-Loop Gaussian Splatting for Detailed and Efficient Surface Reconstruction},
  journal = SIGGRAPH,
  number  = {},
  volume  = {},
  month   = {},
  year    = {2025}
}

@article{Zhang2024ARXIV,
    author = {Zhang, Baowen and Fang, Chuan and Shrestha, Rakesh and Liang, Yixun and Long, Xiao-Xiao and Tan, Ping},
    title = {RaDe-GS: Rasterizing Depth in Gaussian Splatting},
    year = {2026},
    issue_date = {April 2026},
    publisher = {Association for Computing Machinery},
    address = {New York, NY, USA},
    volume = {45},
    number = {2},
    issn = {0730-0301},
    journal = {ACM Trans. Graph.}
}

@ARTICLE{Chen2024ARXIV,
    author={Chen, Danpeng and Li, Hai and Ye, Weicai and Wang, Yifan and Xie, Weijian and Zhai, Shangjin and Wang, Nan and Liu, Haomin and Bao, Hujun and Zhang, Guofeng},
    journal={ IEEE Transactions on Visualization \& Computer Graphics },
    title={{ PGSR: Planar-Based Gaussian Splatting for Efficient and High-Fidelity Surface Reconstruction }},
    year={2025},
    volume={31},
    number={09},
}

@InProceedings{Held_2026_CVPR,
    author    = {Held, Jan and Son, Sanghyun and Vandeghen, Renaud and Rebain, Daniel and Gadelha, Matheus and Zhou, Yi and Cioppa, Anthony and Lin, Ming C. and Van Droogenbroeck, Marc and Tagliasacchi, Andrea},
    title     = {MeshSplatting: Differentiable Rendering with Opaque Meshes},
    booktitle = {Proceedings of the IEEE/CVF Conference on Computer Vision and Pattern Recognition (CVPR)},
    month     = {June},
    year      = {2026},
    pages     = {7320-7329}
}

@InProceedings{Held_2026_3DV,
    author    = {Held, Jan and Vandeghen, Renaud and Deliege, Adrien and Hamdi, Abdullah and Rebain, Daniel and Giancola, Silvio and Cioppa, Anthony and Ghanem, Bernard and Vedaldi, Andrea and Tagliasacchi, Andrea and Van Droogenbroeck, Marc},
    title     = {Triangle Splatting for Real-Time Radiance Field Rendering},
    booktitle = {Proceedings of the International Conference on 3D Vision (3DV)},
    year      = {2026}
}

@article{Laine2020TOG,
  title   = {Modular Primitives for High-Performance Differentiable Rendering},
  author  = {Samuli Laine and Janne Hellsten and Tero Karras and Yeongho Seol and Jaakko Lehtinen and Timo Aila},
  journal = SIGGRAPH,
  year    = {2020},
  volume  = {39},
  number  = {6}
}

@inproceedings{Munkberg2022CVPR,
  title     = {Extracting Triangular 3D Models, Materials, and Lighting From Images},
  author    = {Jacob Munkberg and Wenzheng Chen and Jon Hasselgren and Alex Evans and Tianchang Shen and Thomas M{\"{u}}ller and Jun Gao and Sanja Fidler},
  booktitle = CVPR,
  year      = {2022}
}

@inproceedings{Hasselgren2022NEURIPS,
  title     = {Shape, Light, and Material Decomposition from Images using Monte Carlo Rendering and Denoising},
  author    = {Jon Hasselgren and Nikolai Hofmann and Jacob Munkberg},
  booktitle = NEURIPS,
  year      = {2022}
}

@inproceedings{Fruehauf2024CVPR,
  author    = {Maximilian Fr\"uhauf and Hayko Riemenschneider and Markus Gross and Christopher Schroers},
  title     = {QUADify: Extracting Meshes with Pixel-level Details and Materials from Images},
  booktitle = CVPR,
  year      = {2024}
}

@article{Binninger2025TOG,
  title   = {TetWeave: Isosurface Extraction using On-The-Fly Delaunay Tetrahedral Grids for Gradient-Based Mesh Optimization},
  author  = {Binninger, Alexandre and Wiersma, Ruben and Herholz, Philipp and Sorkine-Hornung, Olga},
  year    ={2025},
  journal = SIGGRAPH,
  month   = {8},
  volume  = {44},
  number  = {4}
}

@inproceedings{Chen2019NEURIPS,
  author    = {Chen, Wenzheng and Gao, Jun and Ling, Huan and Smith, Edward James and Lehtinen, Jaakko and Jacobson, Alec and Fidler, Sanja},
  title     = {Learning to Predict 3D Objects with an Interpolation-based Differentiable Renderer},
  booktitle = NEURIPS,
  year      = {2019}
}

@inproceedings{Liu2019ICCV,
  author    = {Liu, Shichen and Li, Tianye and Chen, Weikai and Li, Hao},
  title     = {Soft Rasterizer: A Differentiable Renderer for Image-Based 3D Reasoning},
  booktitle = ICCV,
  year      = {2019}
}

@article{Jatavallabhula2019ARXIV,
  author  = {Jatavallabhula, Krishna Murthy and Smith, Edward and Lafleche, Jean-Francois and Fuji Tsang, Clement and Rozantsev, Artem and Chen, Wenzheng and Xiang, Tommy and Lebaredian, Rev and Fidler, Sanja},
  title   = {Kaolin: A PyTorch Library for Accelerating 3D Deep Learning Research},
  journal = {arXiv.org},
  volume  = {1911.05063},
  year    = {2019}
}

@article{Ravi2020ARXIV,
  author  = {Ravi, Nikhila and Reizenstein, Jeremy and Novotny, David and Gordon, Taylor and Lo, Wan-Yen and Johnson, Justin and Gkioxari, Georgia},
  title   = {Accelerating 3D Deep Learning with PyTorch3D},
  journal = {arXiv.org},
  volume  = {2007.08501},
  year    = {2020}
}

@inproceedings{Chen2021NEURIPS,
  author    = {Chen, Wenzheng and Litalien, Joey and Gao, Jun and Wang, Zian and Fuji Tsang, Clement and Khamis, Sameh and Litany, Or and Fidler, Sanja},
  title     = {{DIB-R++}: Learning to Predict Lighting and Material with a Hybrid Differentiable Renderer},
  booktitle = NEURIPS,
  year      = {2021}
}

@article{Shen2023TOG,
  author  = {Shen, Tianchang and Munkberg, Jacob and Hasselgren, Jon and Yin, Kangxue and Wang, Zian and Chen, Wenzheng and Gojcic, Zan and Fidler, Sanja and Sharp, Nicholas and Gao, Jun},
  title   = {Flexible Isosurface Extraction for Gradient-Based Mesh Optimization},
  journal = SIGGRAPH,
  volume  = {42},
  number  = {4},
  year    = {2023}
}

@inproceedings{Peng2021NEURIPS,
  author         = {Songyou Peng and Chiyu Max Jiang and Yiyi Liao and Michael Niemeyer and Marc Pollefeys and Andreas Geiger},
  title          = {Shape As Points: A Differentiable Poisson Solver},
  booktitle      = NEURIPS,
  year           = {2021}
}

@article{Chen2023ARXIVb,
  Title   = {Factor Fields: A Unified Framework for Neural Fields and Beyond},
  Author  = {Chen, Anpei and Xu, Zexiang and Wei, Xinyue and 
  		     Tang, Siyu and Su, Hao and Geiger, Andreas},
  Journal   = {arXiv.org},
  Year      = {2023},
  Volume    = {2302.01226}
}

@article{Chen2023TOG,
  Title   = {Dictionary Fields: Learning a Neural Basis Decomposition},
  Author  = {Chen, Anpei and Xu, Zexiang and Wei, Xinyue and 
  		   Tang, Siyu and Su, Hao and Geiger, Andreas},
  Journal = SIGGRAPH,
  Year    = {2023},
  Number  = {4},
  Pages   = {1--12},
  Volume  = {42}
}

@article{adaptiveshells2023,
  author = {Zian Wang and Tianchang Shen and Merlin Nimier-David and Nicholas Sharp and Jun Gao and Alexander Keller and Sanja Fidler and Thomas M\"uller and Zan Gojcic},
  title = {Adaptive Shells for Efficient Neural Radiance Field Rendering},
  journal = {ACM Trans. Graph.},
  issue_date = {December 2023},
  volume = {42},
  number = {6},
  year = {2023},
  articleno = {259},
  publisher = {ACM},
  address = {New York, NY, USA}
}

@inproceedings{Esposito2025VolSurfs,
  author    = {Esposito, Stefano and Chen, Anpei and Reiser, Christian and Rota Bulò, Samuel and Porzi, Lorenzo and Schwarz, Katja and Richardt, Christian and Zollhoefer, Michael and Kontschieder, Peter and Geiger, Andreas},
  title     = {Volumetric Surfaces: Representing Fuzzy Geometries with Layered Meshes},
  booktitle = {IEEE/CVF Conference on Computer Vision and Pattern Recognition (CVPR)},
  year={2025}
}

@article{guedon2024frosting,
 title={Gaussian Frosting: Editable Complex Radiance Fields with Real-Time Rendering},
 author={Gu{\'e}don, Antoine and Lepetit, Vincent},
 journal={ECCV},
 year={2024}
}

@article{Yang2025IMLSSplattingEM,
  title={IMLS-Splatting: Efficient Mesh Reconstruction from Multi-view Images via Point Representation},
  author={Kaizhi Yang and Liu Dai and Isabella Liu and Xiaoshuai Zhang and Xiaoyan Sun and Xuejin Chen and Zexiang Xu and Hao Su},
  journal={ACM Transactions on Graphics (TOG)},
  year={2025},
  volume={44},
  pages={1 - 11},
  url={https://api.semanticscholar.org/CorpusID:280537344}
}

@inproceedings{
    zhang2026mesh,
    title={Mesh Splatting for End-to-end Multiview Surface Reconstruction},
    author={Ruiqi Zhang and JiachengWU and Jie Chen},
    booktitle={The Fourteenth International Conference on Learning Representations},
    year={2026},
    url={https://openreview.net/forum?id=PSgps4JXTb}
}

@article{quadfields,
  author    = {Gopal Sharma and Daniel Rebain and Andrea Tagliasacchi and Kwang Moo Yi},
  title     = {Volumetric Rendering with Baked Quadrature Fields},
  journal   = {ECCV},
  year      = {2024},
}

@article{Reiser2024SIGGRAPH,
    title={Binary Opacity Grids: Capturing Fine Geometric Detail for Mesh-Based View Synthesis},
    author={Christian Reiser and Stephan Garbin and Pratul P. Srinivasan and 
        Dor Verbin and Richard Szeliski and Ben Mildenhall and Jonathan T. Barron and Peter Hedman
        and Andreas Geiger},
    journal={SIGGRAPH},
    year={2024}
}

@inproceedings{Jensen2014CVPR,
  author    = {Jensen, Rasmus and Dahl, Anders and Vogiatzis, George and Tola, Engin and Aan{\ae}s, Henrik},
  title     = {Large Scale Multi-view Stereopsis Evaluation},
  booktitle = CVPR,
  year      = {2014}
}

@inproceedings{Kuang2023NEURIPS,
  author    = {Kuang, Zhengfei and Zhang, Yunzhi and Yu, Hong-Xing and Agarwala, Samir and Wu, Shangzhe and Wu, Jiajun},
  title     = {Stanford-ORB: A Real-World 3D Object Inverse Rendering Benchmark},
  booktitle = NEURIPS,
  year      = {2023}
}

@article{Knapitsch2017TOG,
  author  = {Knapitsch, Arno and Park, Jaesik and Zhou, Qian-Yi and Koltun, Vladlen},
  title   = {Tanks and Temples: Benchmarking Large-Scale Scene Reconstruction},
  journal = SIGGRAPH,
  volume  = {36},
  number  = {4},
  year    = {2017}
}

@article{yao2020blendedmvs,
  title={BlendedMVS: A Large-scale Dataset for Generalized Multi-view Stereo Networks},
  author={Yao, Yao and Luo, Zixin and Li, Shiwei and Zhang, Jingyang and Ren, Yufan and Zhou, Lei and Fang, Tian and Quan, Long},
  journal={Computer Vision and Pattern Recognition (CVPR)},
  year={2020}
}

@inproceedings{Wang2024NEURIPS,
  author    = {Wang, Fangjinhua and Rakotosaona, Marie-Julie and Niemeyer, Michael and Szeliski, Richard and Pollefeys, Marc and Tombari, Federico},
  title     = {UniSDF: Unifying Neural Representations for High-Fidelity 3D Reconstruction of Complex Scenes with Reflections},
  booktitle = NEURIPS,
  year      = {2024}
}

@article{marching_tetrahedra,
    author={Akio Doi and Akio Koide},
    journal={IEICE TRANSACTIONS on Information},
    title={An Efficient Method of Triangulating Equi-Valued Surfaces by Using Tetrahedral Cells},
    year={1991},
    volume={E74-D},
    number={1},
    pages={214-224},
    doi={},
    ISSN={},
    month={January},}

@article{delaunay1934sphere,
  title={Sur la sph{\`e}re vide},
  author={Delaunay, Boris},
  journal={Bulletin de l'Acad{\'e}mie des Sciences de l'URSS, Classe des sciences math{\'e}matiques et naturelles},
  volume={6},
  pages={793--800},
  year={1934}
}

@incollection{lee1980two,
  title={Two algorithms for constructing a Delaunay triangulation},
  author={Lee, Der-Tsai and Schachter, Bruce J},
  booktitle={International Journal of Computer and Information Sciences},
  volume={9},
  number={3},
  pages={219--242},
  year={1980},
  publisher={Springer}
}

@inproceedings{adamw,
  author       = {Ilya Loshchilov and
                  Frank Hutter},
  title        = {Decoupled Weight Decay Regularization},
  booktitle    = {7th International Conference on Learning Representations, {ICLR} 2019,
                  New Orleans, LA, USA, May 6-9, 2019},
  publisher    = {OpenReview.net},
  year         = {2019},
  url          = {https://openreview.net/forum?id=Bkg6RiCqY7},
  bibsource    = {dblp computer science bibliography, https://dblp.org}
}

@misc{instant-nsr-pl,
    Author = {Yuan-Chen Guo},
    Year = {2022},
    Note = {https://github.com/bennyguo/instant-nsr-pl},
    Title = {Instant Neural Surface Reconstruction}
}

\section*{}
\clearpage

\appendix
\section{Implementation Details}
\label{sec:A_implementation_details}

Our implementation builds upon the Instant-NSR-PL~\cite{instant-nsr-pl} codebase, which combines NeuS with multiresolution hash encodings as introduced in Instant-NGP~\cite{Mueller2022TOG}, and adds full support for PyTorch Lightning. We normalize each scene’s coordinate system such that it is centered around the approximate intersection point of all camera viewing directions and scaled so that the distance between the closest camera and the scene center equals 1. The tetrahedral grid and hash-grid encoder are both initialized at the scene center with a radius of 0.6.

We encode both the SDF and the local appearance function in a single hashgrid consisting of 16 levels, with a minimum resolution of 32 and a maximum resolution of 2048. Its feature MLP contains one hidden layer of 64 neurons and employs weight normalization. The color network is an MLP with two hidden layers of 64 neurons and uses 4th-degree spherical harmonics to encode normal directions.

We initialize the tetrahedral grid with 250k points sampled uniformly within the region of interest, and the SDF is initialized as a sphere with a radius of 0.5. The representation is trained jointly with volumetric and mesh-based rendering for 3000 iterations, followed by 1500 iterations using only mesh-based rendering. We anneal the mesh-based loss from $0.5$ to $1$ and the volumetric loss from $0.5$ to $0$ between iterations 1000 and 3000. Grid densification starts at iteration 2000, where 50\% of the tetrahedra are subdivided every 200 iterations until reaching a maximum of 1M points.

During training, each batch uses between 1 and 3 samples for the offset sampling procedure, where the number of samples is drawn uniformly at random per batch. We find that mixing different sample counts yields more stable convergence. 
We optimize with batches of 4 input images and an adaptive number of rays during volumetric training as well as gradient accumulation across 4 batches. Optimization is performed using the AdamW optimizer~\cite{adamw} (Adam with decoupled weight decay) with $\beta = [0.9, 0.99]$, a learning rate of $1\text{e-}4$ for tetrahedral vertex positions, and $1\text{e-}2$ for all other parameters. For our final model, we set $\lambda_{\text{mask}} = 1.0,$ and  $\lambda_{\text{photo}} = 10.0.$
All experiments for our method were conducted on an NVIDIA H200 GPU with 141 GB of memory.

\subsection{Volumetric Rendering Details}
\label{subsec:volumetric_rendering_details}
To facilitate faster rendering and avoid evaluating samples in empty space, we employ a multiresolution occupancy grid. This grid is updated every $16$ iterations based on the signed distance function (SDF) values of random samples drawn within each voxel. For ray marching, we evaluate at most $1280$ samples per ray. We utilize a dynamic batch size to maximize GPU utilization, adjusting the number of rays per batch based on the average number of volumetric samples required per ray in previous iterations. We begin with a batch size of $256$ rays and cap the maximum at $8192$ rays.

\textbf{Loss Terms.} Similar to the mesh-based rendering branch, we supervise the volumetric branch with a photometric loss $\mathcal{L}^{vol}_{photo}$ and a mask loss $\mathcal{L}^{vol}_{mask}$ (see Equations \ref{eq:photometric_loss} and \ref{eq:mask_loss} in the main text). In addition, we employ an Eikonal regularization term $\mathcal{L}_{eik}$ to encourage the learned field to approximate a valid signed distance function. This is defined as the $\ell_1$-error between the gradient norm of the SDF samples and unity:

\begin{equation}
\mathcal{L}_{eik} = \frac{1}{|\mathcal{S}|} \sum_{\mathbf{x} \in \mathcal{S}} \left( \| \nabla f(\mathbf{x}) \|_2 - 1 \right)^2,
\end{equation}

where $\mathcal{S}$ represents the set of sampled points and $f(\mathbf{x})$ denotes the SDF value at point $\mathbf{x}$. The gradient of the SDF for each sample is derived analytically. We find that Eikonal regularization has a significantly positive impact, particularly for scenes captured in-the-wild with complicated lighting conditions. The full volumetric loss function is given by:

\begin{equation}
\mathcal{L}_{vol} = \lambda_{photo} \mathcal{L}^{vol}_{photo} + \lambda_{mask} \mathcal{L}^{vol}_{mask} + \lambda_{eik} \mathcal{L}_{eik},
\end{equation}

where $\lambda_{photo}$, $\lambda_{mask}$, and $\lambda_{eik}$ are weighting parameters controlling the strength of the individual terms. We set $\lambda_{photo} = 10.0$, $\lambda_{mask} = 1.0$, and $\lambda_{eik} = 0.1$.

The total loss function for the joint optimization is defined as follows:

\begin{equation}
\mathcal{L} = \lambda_{mesh} \mathcal{L}_{mesh} + \lambda_{vol} \mathcal{L}_{vol}.
\end{equation}

To transition from the hybrid initialization to the final mesh-based optimization, we employ a linear annealing schedule. We initialize both $\lambda_{mesh}$ and $\lambda_{vol}$ to $0.5$. Between iterations $1000$ and $3000$, we linearly anneal $\lambda_{mesh}$ to $1.0$ and $\lambda_{vol}$ to $0.0$, effectively fading out the volumetric branch once the coarse geometry is established.

\subsection{Depth Offset Sampling}
\begin{figure}[h]
  \centering
   \includegraphics[width=\linewidth]{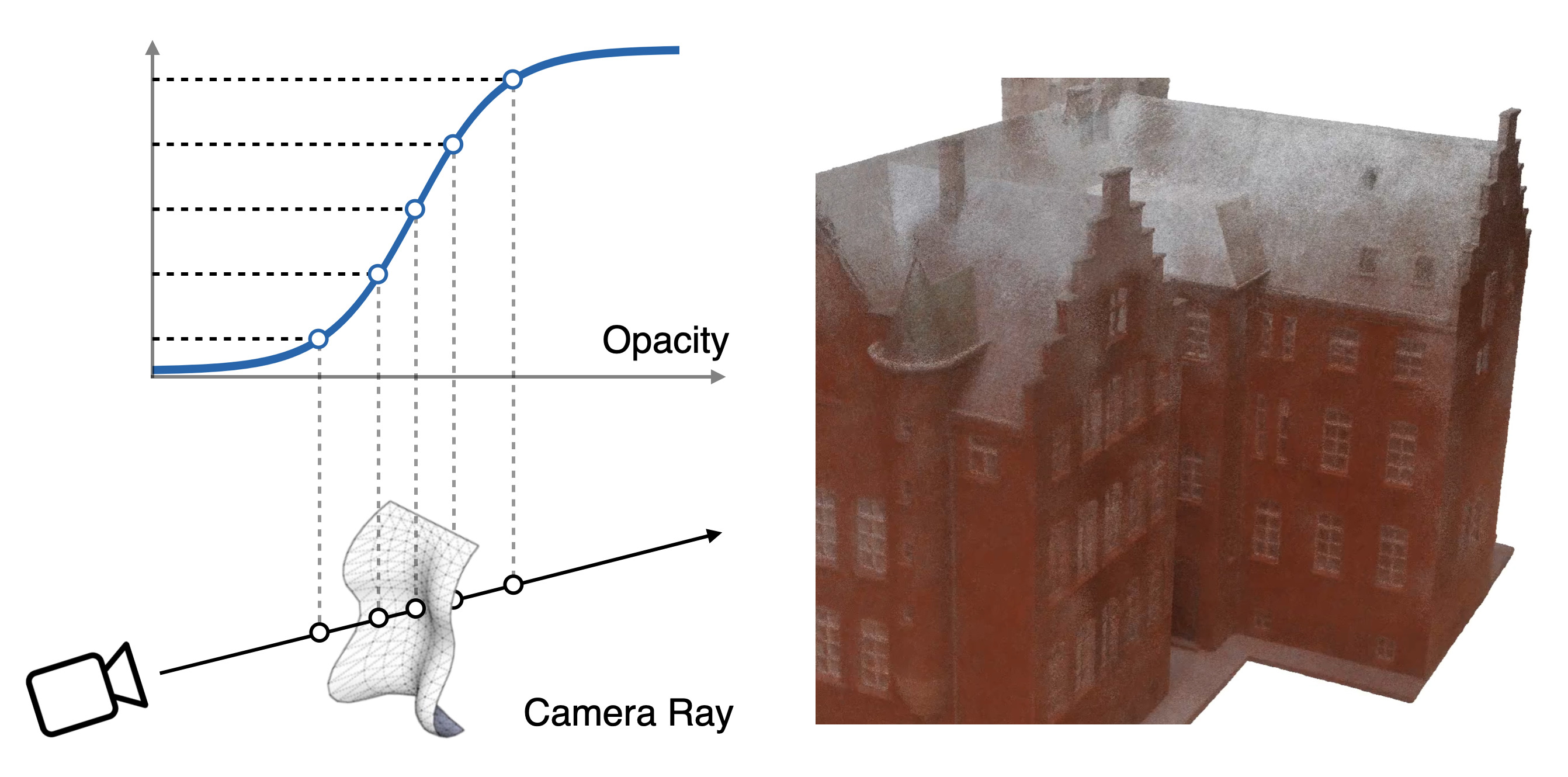}
    \caption{\textbf{Depth-offset sampling.} (Left) We sample multiple points along each camera ray by performing inverse sampling with respect to the opacity function $\Psi(t)$. The sigmoid shape of $\Psi$ concentrates samples near the mesh surface, where the SDF transitions through zero. (Right) early stage reconstruction rendered using sampled depth offsets. }
   \label{fig:bandwidth_sampling}
\end{figure}

In the mesh-based rendering path, we sample multiple depth values $d_i$ around the rendered depth $d$ for each pixel where individual depth samples are computed as follows. We assume that the SDF is locally planar, oriented in the direction of the mesh normal. Let $\mathbf{o}$ denote the camera origin and $t_0$ the intersection point of the ray and the mesh under the ray parameterization $\mathbf{o}+ t \mathbf{d}$. The SDF $f(t)$ along the ray is then linear
\[
f(t) = (\mathbf{n}\cdot\mathbf{d})(t-t_0).
\]
Note that we consider the camera to lie outside the observed surface; hence, the SDF decreases along the ray. Similar to NeuS \cite{Wang2021NEURIPS}, we assume that the opacity along the ray follows a sigmoid curve centered at the intersection point, that is,
\[
\Psi(t) = \frac{1}{1+e^{\sigma f(t)}},
\]
see Figure~\ref{fig:bandwidth_sampling} for an illustration. Here, $\sigma$ is a learnable parameter that controls the steepness of the opacity function. To draw $n$ depth samples, we perform inverse sampling with respect to $\Psi$. Specifically, we divide the range of the opacity function, $[0,1]$, into $n$ intervals with corresponding midpoints
\[
[T_i, T_{i+1}], \qquad \alpha_i = \frac{T_i+T_{i+1}}{2}, \qquad i \in\{0,\dots,n\},
\]
where $T_0 = 0$, $T_{n+1}=1$, and the remaining $T_i$ are sampled via stratified sampling, \ie, $T_i \sim U\left( \frac{i-1}{n}, \frac{i}{n} \right)$. 
Using the inverse transform of $\Psi$ given by
\[
\Psi^{-1}(\alpha)
= f^{-1}\!\left(\frac{\log\!\left(\frac{\alpha}{1-\alpha}\right)}{\sigma}\right)
= t_0 + \frac{\log\!\left(\frac{1-\alpha}{\alpha}\right)}{(\mathbf{n}\cdot\mathbf{d})\,\sigma},
\]
we calculate the 3D positions along the ray corresponding to the sampled interval midpoints
\[
\mathbf{x}_i = \mathbf{o} + \Psi^{-1}(\alpha_i)\mathbf{d}.
\]
For each sampled position $\mathbf{x}_i$, its corresponding color $c_i$ is computed via Eq.~\ref{eqn:color_network}. The final pixel color is then obtained as a weighted average over all sampled colors according to their contribution to the ray opacity
\begin{equation}
C = \sum_i \left(T_{i+1}-T_i\right) c_i.
\end{equation}

\section{Large-Scale Reconstruction}
\label{sec:large_scale_reconstruction}

In this section, we \rev{discuss details of the initialization and implementation in large-scale reconstruction scenarios.} 

% In this section, we demonstrate the applicability of our method to large-scale reconstruction scenarios. For unbounded, large-scale scenes, we observe that the spherical initialization employed in the object-centric case is not sufficient. We instead propose initializing the optimization from a coarse SDF estimate and learning a residual field relative to this base geometry.

\subsection{Initial SDF}
\begin{figure}[h]
  \centering
   \includegraphics[width=\linewidth]{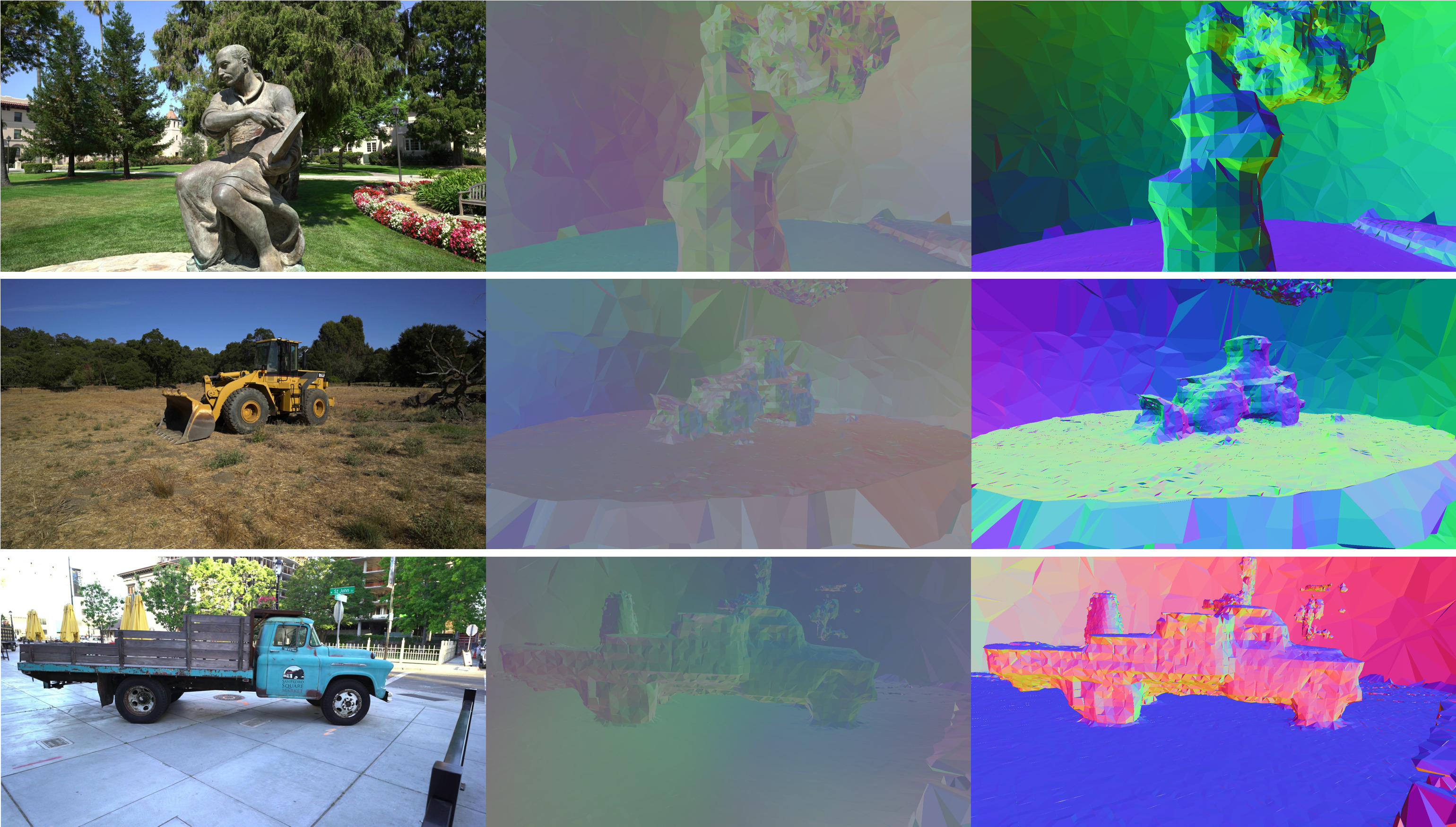}
   \caption{\textbf{Initialization Examples.} Visualizations of the coarse initial meshes derived from VGGT depth priors, which serve as the base geometry for our residual learning framework.}
   \label{fig:tnt_initialization}
\end{figure}

We leverage strong priors from large-scale pretrained models to obtain a coarse preliminary mesh estimate. Specifically, we utilize VGGT~\cite{Wang2025CVPR} to predict camera poses and per-view depth maps. These predictions are subsequently employed to extract an initial mesh via TSDF fusion. Figure~\ref{fig:tnt_initialization} illustrates examples of these initial meshes for scenes from the Tanks and Temples (TnT) dataset~\cite{Knapitsch2017TOG}.

This initial mesh is distilled into a discretized SDF, denoted as $s_0$, stored on a static 3D grid by computing the exact SDF value for each grid vertex. During optimization, the query for the SDF value at a point $\mathbf{x}$ is composed of the trilinear interpolation of the base grid $s_0$ and a learnable residual $s_{res}$ parameterized by a hashgrid:
\begin{equation}
    s(\mathbf{x}) = s_0(\mathbf{x}) + \lambda_{res}s_{res}(\mathbf{x}),
\end{equation}
where $\lambda_{res}$ is a hyperparameter governing the magnitude of the learned residual.
We restrict explicit mesh reconstruction to a predetermined region of interest (ROI) centered on the scene. To model the background environment outside this region, we employ a separate environment light, represented as a 2D hash grid and a shallow decoder MLP. This network takes as input the viewing direction, mapped to $[0,1]^2$ via spherical coordinates, and outputs the corresponding background color.

\subsection{Appearance Model}
Large-scale scenes often exhibit complex lighting conditions, rendering our single-MLP color model prone to geometric artifacts. To mitigate this, we adopt the appearance model from UniSDF~\cite{Wang2024NEURIPS}, which decomposes outgoing radiance into view-dependent and reflected components.

For a sample along a camera ray at position $\mathbf{x}$, let $\mathbf{d}$ denote the viewing direction, $\mathbf{n}$ the surface normal, and $\boldsymbol{\xi}$ the appearance vector obtained from the appearance hash grid. We compute the reflected viewing direction $\boldsymbol{\omega}_r$ as:
\begin{equation}
    \boldsymbol{\omega}_r = \mathbf{d} - 2(\mathbf{d} \cdot \mathbf{n})\mathbf{n}.
\end{equation}
The architecture consists of a blending weight MLP $f_{w}$ and two distinct radiance field MLPs: $f_{cam}$, which models the view-dependent component, and $f_{ref}$, which models the reflection-dependent component. The components are computed as follows:
\begin{align}
    \mathbf{c}_{cam} &= f_{cam}(\mathbf{x}, \mathbf{d}, \mathbf{n}, \boldsymbol{\xi}) \\
    \mathbf{c}_{ref} &= f_{ref}(\mathbf{x}, \boldsymbol{\omega}_r, \mathbf{n}, \boldsymbol{\xi}) \\
    w &= \sigma(f_w(\mathbf{x}, \mathbf{n}, \boldsymbol{\xi})),
\end{align}
where $\sigma(\cdot)$ denotes the sigmoid function. The final color $\mathbf{c}$ is obtained via linear blending:
\begin{equation}
    \mathbf{c} = w \mathbf{c}_{ref} + (1-w) \mathbf{c}_{cam}.
\end{equation}
Differing from the original formulation described in UniSDF, this blending is performed at the sample level, prior to volume rendering integration. Furthermore, to account for exposure variations inherent in the TnT dataset, we optimize per-image appearance embeddings following \cite{MartinBrualla2021CVPR}.

\subsection{Implementation Details}

\textbf{Initialization.} To extract the initial coarse geometry, we filter depth predictions from VGGT, retaining only the top 50\% of the most confident pixels. The remaining depth estimates are integrated using TSDF fusion with a voxel size of $0.02$. The resulting mesh is discretized into a 3D grid with a resolution of $64^3$ to form the base SDF $s_0$. The weights of the residual hashgrid are randomly initialized, and the residual scaling factor is set to $\lambda_{res} = 0.1$. 

\textbf{Optimization and Scheduling.} We define the ROI as a cubic volume with a side length of $1.5 \times \bar{d}$, where $\bar{d}$ is the median distance from the cameras to the scene center (computed as the mean camera position). We extend the optimization to 10,000 iterations. Accordingly, the densification schedule is adapted to run from iteration 4,000 to 8,000, with a minimum interval of 400 steps between densification events. Pruning is performed every 1,000 iterations, starting from iteration 4,000. For hybrid training, the weights $\lambda_{\text{mesh}}$ and $\lambda_{\text{vol}}$ are initialized to 0.5. Between iterations 4,000 and 8,000, we linearly anneal $\lambda_{\text{mesh}}$ to 1.0 and $\lambda_{\text{vol}}$ to 0.0.

\textbf{Network Architecture.} The color MLPs $f_{ref}$ and $f_{cam}$ comprise two hidden layers with 64 neurons each, while the weight MLP $f_w$ contains a single hidden layer of 64 neurons. The background model utilizes a multi-resolution hash grid with 12 levels, a minimum resolution of 16, a maximum resolution of 338, and 2 features per level, followed by a single-layer MLP with 64 neurons.

\textbf{Hyperparameters.} We use a learning rate of $5 \times 10^{-4}$ for grid positions and $0.05$ for the background network parameters. All other network parameters are optimized with a learning rate of $0.005$. A linear learning rate warmup is applied over the first 1,000 steps. Unless otherwise specified, all other parameters match those described in the main methodology. 

\section{Mesh Regularization}
\label{sec:D_mesh_regularization}

\begin{figure}[t]
  \centering
  \includegraphics[width=1.0\linewidth]{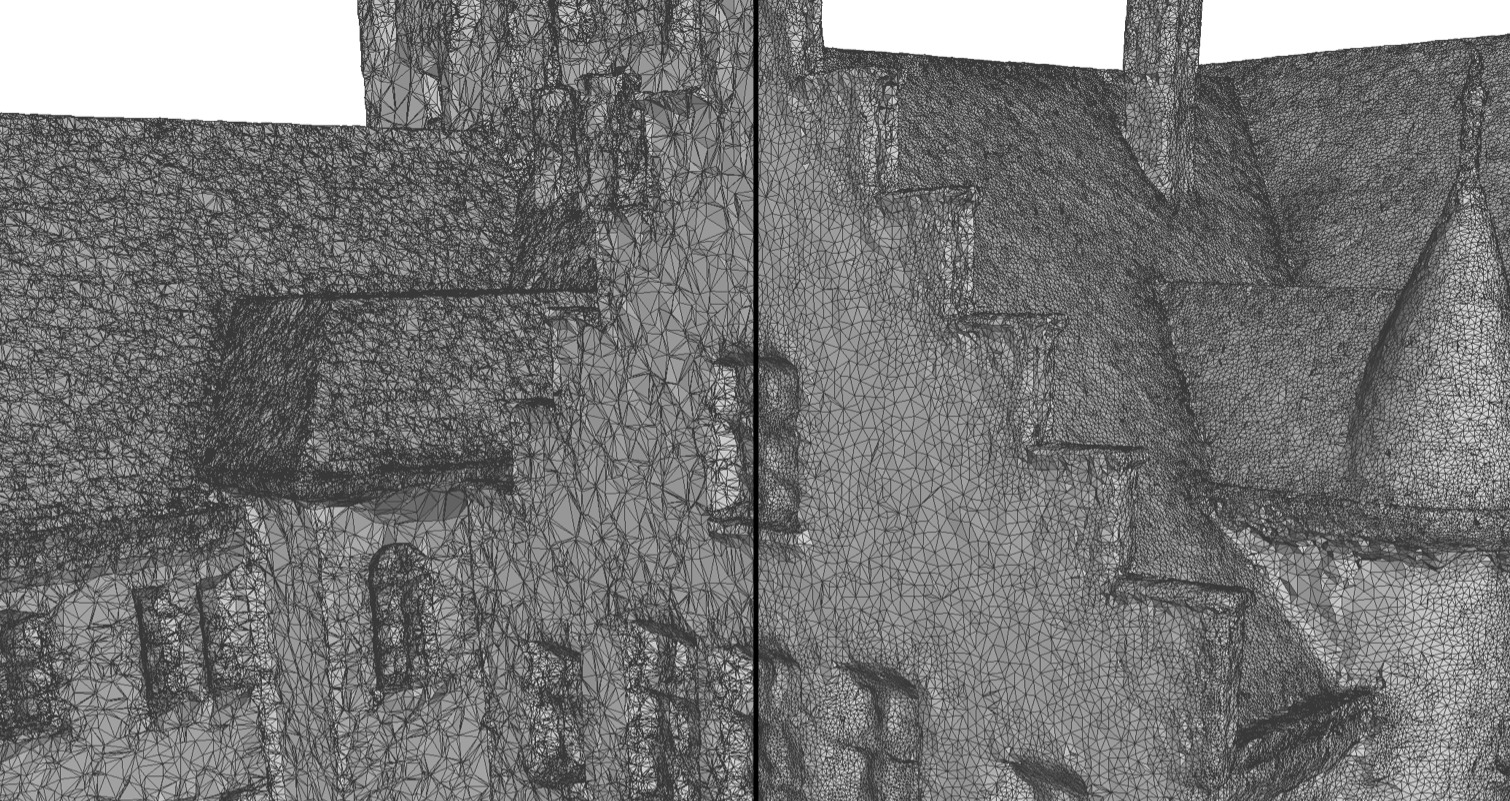}
    \caption{Incorporating a triangle regularization term that penalizes deviations from regular triangles in the output mesh leads to a more evenly tesselated reconstructed surface (right) than our base method (left).}
   \label{fig:mesh_regularization}
\end{figure}

 A significant advantage of optimizing a mesh based representation in an end-to-end fashion is the ability to directly regularize the mesh based on desirable properties. We showcase this advantage by implementing a regularization term, previously introduced in \cite{Binninger2025TOG}, which penalizes angles in the output mesh that deviate from $\frac{\pi}{3}$ to promote equilateral triangles:
\[
\mathcal{L}_{\text{fairness}} = \sum_{f \in F} \frac{1}{3} \sum_{i=1}^{3} \left( \theta^f_i - \frac{\pi}{3} \right)^2,
\]
where $F$ denotes the set of all mesh faces, and $\theta^f_i$ represents the interior angles of face $f$. \autoref{fig:mesh_regularization} shows the effect of this regularizer on the reconstructed mesh under a regularization strength of $\lambda_{fairness} = 0.01$. We find that introducing this fairness term comes at a slight disadvantage to visual and geometric accuracy; therefore, we do not include it in the evaluation of our main method.

\section{DTU Evaluation Bias}
\label{sec:C_DTU_dataset_bias}

\begin{figure}[t]
  \centering
   \includegraphics[width=0.85\linewidth]{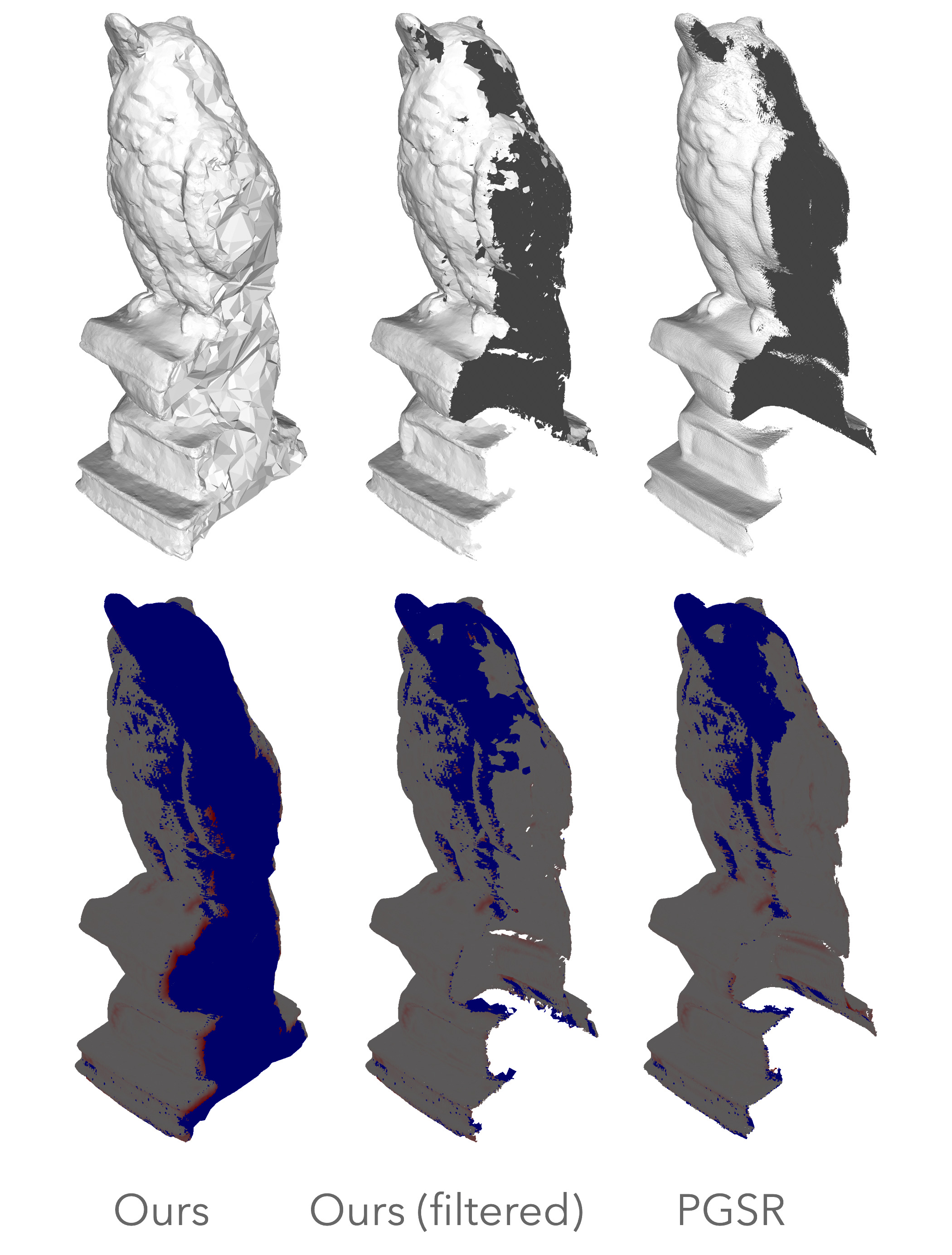}
   \caption{Reconstructed meshes and corresponding error maps for a scene from the DTU dataset. Discarded points are colored in blue, while evaluated points are shown in grey for low error and red for high error. Our method produces closed meshes, whereas PGSR reconstructs only a front-facing surface. Notably, points reconstructed by our method near the mesh boundaries exhibit significant error contributions. This problem is largely resolved by our filtering proecdure.}\label{fig:dtu_bias}
\end{figure}

During evaluation of the DTU dataset, we observed that the standard evaluation protocols exhibit a bias toward methods that produce incomplete or open-surface meshes. This arises from the fact that the ground truth point clouds represent only the front-facing surfaces of objects, leaving the backsides and occluded regions undefined. Our method, which recovers a closed, watertight  mesh, also reconstructs unseen back-facing surfaces. In contrast, TSDF-fusion-based methods, such as PGSR, generally extract only the visible front-facing surface, resulting in an open mesh that inherently overlaps better with the partial ground truth. Points sampled from unobserved or sparsely observed reconstructed back-facing surfaces often lack corresponding nearest neighbors in the incomplete ground truth point clouds. Although these regions are geometrically valid under a watertight reconstruction assumption, they nevertheless contribute significantly to the overall error term, see \autoref{fig:dtu_bias}.
To ensure a fair comparison, we apply a simple visibility-based filtering procedure that removes triangles in sparsely observed regions before evaluation. Specifically, we render each reconstructed mesh at twice the input resolution from every training view and mark any triangle visible in at most one view as unobserved. We then remove all vertices whose incident triangles are exclusively unobserved. The resulting mesh preserves the front-facing surface almost entirely while discarding the back-facing regions that the ground truth does not cover, as illustrated in \autoref{fig:dtu_bias}. We emphasize that this filtering does not improve the geometric quality of the reconstruction itself, and is applied solely to align the support of our prediction with that of the ground truth so that the comparison against baseline methods is not skewed by the choice of surface representation. Beyond the standard geometric evaluations, we further encourage taking into account visual comparisons of the extracted meshes.

\section{Memory Analysis}
\label{sec:E_memory_analysis}

We conduct an extensive analysis of peak GPU memory consumption (VRAM) and training runtime on the DTU dataset (Tab.~\ref{tab:memory_benchmark}). Our full method requires approximately 32 GB of peak GPU memory, allowing training to fit within a high-end consumer or workstation GPU. In addition, we evaluate a more memory-efficient variant of our approach that achieves only slightly reduced reconstruction quality while simultaneously reducing training time and lowering the memory footprint to below 24 GB. This variant uses a batch size of a single image together with gradient accumulation over four optimization steps, thereby significantly decreasing the instantaneous memory requirements. A substantial portion of the current memory consumption originates from the RadFoam-based Delaunay triangulation procedure. Replacing this component with a CPU-based triangulation implementation could further reduce GPU memory usage, albeit at the cost of increased runtime. We further note that we were unable to reproduce the runtimes reported in MILo when re-running their publicly available codebase, consistently observing substantially longer training times instead. Consequently, the runtimes reported in our memory benchmark differ from both the values stated in the main paper and those originally reported by the authors.

\begin{table}[t]
\centering
\caption{Comparison of memory usage, runtime, and reconstruction quality on DTU. Peak VRAM is measured during training; runtime is the average wall-clock training time per scene; Chamfer distance is the mean over all evaluated
scenes (lower is better).}
\label{tab:memory_benchmark}
\resizebox{\columnwidth}{!}{
\begin{tabular}{lcccc}
\toprule
Method & \makecell{Avg.\ Peak\\VRAM (GB)} & \makecell{Max.\ Peak\\VRAM (GB)} & \makecell{Avg.\ Runtime\\(min)} & \makecell{Chamfer\\Distance $\downarrow$} \\
\midrule
PGSR                     & 3.9   & 5.5   & 41.4  & 0.53 \\
MiLO                     & 5.2   & 8.9   & 69.6  & 0.68 \\
\midrule
ADELE (full)         & 31.9  & 33.1 & 30.3 & 0.53 \\
ADELE (smaller batch) & 21.4 & 22.5 & 14 & 0.56 \\
\bottomrule
\end{tabular}
}
\end{table}
\section{Additional Results}
\label{sec:F_additional results}

\subsection{Stanford ORB }
\begin{figure*}[t]
  \centering
  \includegraphics[width=1.0\linewidth]{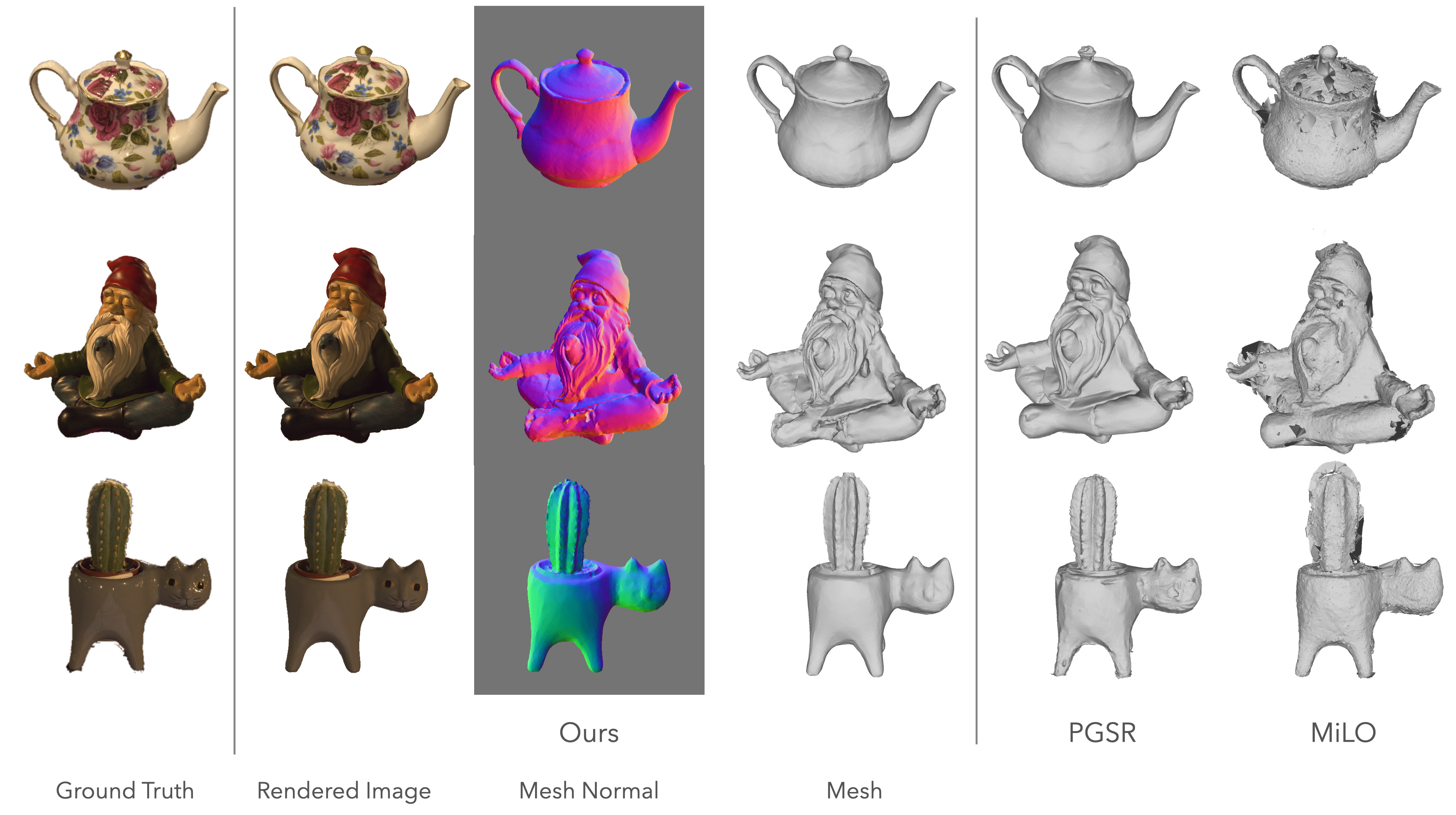}
    \caption{\textbf{Qualitative comparison on the Stanford ORB dataset.} For each scene, we show the ground truth image, our rendered image, our predicted mesh normals, and mesh reconstructions from our method, PGSR, and MiLO. Our method produces detailed meshes that faithfully capture object silhouettes and surface+details. PGSR struggles with reflective surfaces, while MiLO tends to produce simplified geometry that misses fine-grained features such as the gnome's beard and the cactus spines.}
   \label{fig:orb_results}
\end{figure*}

In this section we report additional geometric reconstruction results on the Stanford ORB dataset. For all methods evaluated on this dataset, we perform a culling step on the reconstructed meshes, removing any triangles that are not visible from at least one training view. This post-processing step ensures that the evaluation focuses on surface geometry actually supervised by the input images and does not penalize spurious interior geometry. 
Table~\ref{tab:orb_results} presents the Chamfer Distance metrics for the selected Stanford ORB scenes. The results indicate that all three methods achieve comparable geometric accuracy across the majority of the dataset with the exception of some outliers.  Qualitatively (Fig.~\ref{fig:orb_results}), PGSR exhibits severe artifacts on the highly reflective pitcher scene, while MiLo does not capture the object silhouette well. In contrast, our method achieves detailed reconstruction while avoiding the aforementioned artifacts but exhibits artifacts on the rough reflective surface of the gnome scene.

\begin{table}[h]
\caption{Quantitative comparison on Stanford ORB. We report Chamfer distance ($\downarrow$) scaled by $2e3$ as proposed in the official evaluation guidelines, and triangle count of the extracted meshes.}
\label{tab:orb_results}
\centering
\small
%\resizebox{\linewidth}{!}{ % Uncomment this line if the table is too wide for your page
\begin{tabular}{l|cccccc}
\multicolumn{1}{c}{} & \multicolumn{6}{c}{Chamfer Distance $ \times $ $  2e3\downarrow$} \\
Scene & Gnome & Cactus & Ball & Teapot & Pitcher &Avg \\
\hline
PGSR & \textbf{0.21} & 0.08 &0.07  & 0.16 & 2.10 & 0.52 \\
MiLO & \textbf{0.21} & 0.12 & 0.11 & 0.21 & \textbf{0.90} &  \textbf{0.31} \\
Ours & 0.42 & \textbf{0.05} & \textbf{0.05} & \textbf{0.14} & 0.99 & 0.33 \\
\multicolumn{7}{c}{} \\ % Empty row for spacing
\multicolumn{1}{c}{} & \multicolumn{6}{c}{Triangle Count$\downarrow$} \\
Scene & Gnome & Cactus & Ball & Teapot & Pitcher &Avg \\
\hline
PGSR & 890k & 420k & 756k & 485k & 631k & 636k\\
MiLO & 427k & 214k & 336k & 386k & 105k &   294k\\
Ours & 655k & 309k & 662k & 439k & 414k & 496k \\
\end{tabular}
%} % Uncomment if using resizebox
\end{table}

\subsection{Novel View Synthesis}
\begin{figure*}[t]
  \centering
   \includegraphics[width=\linewidth]{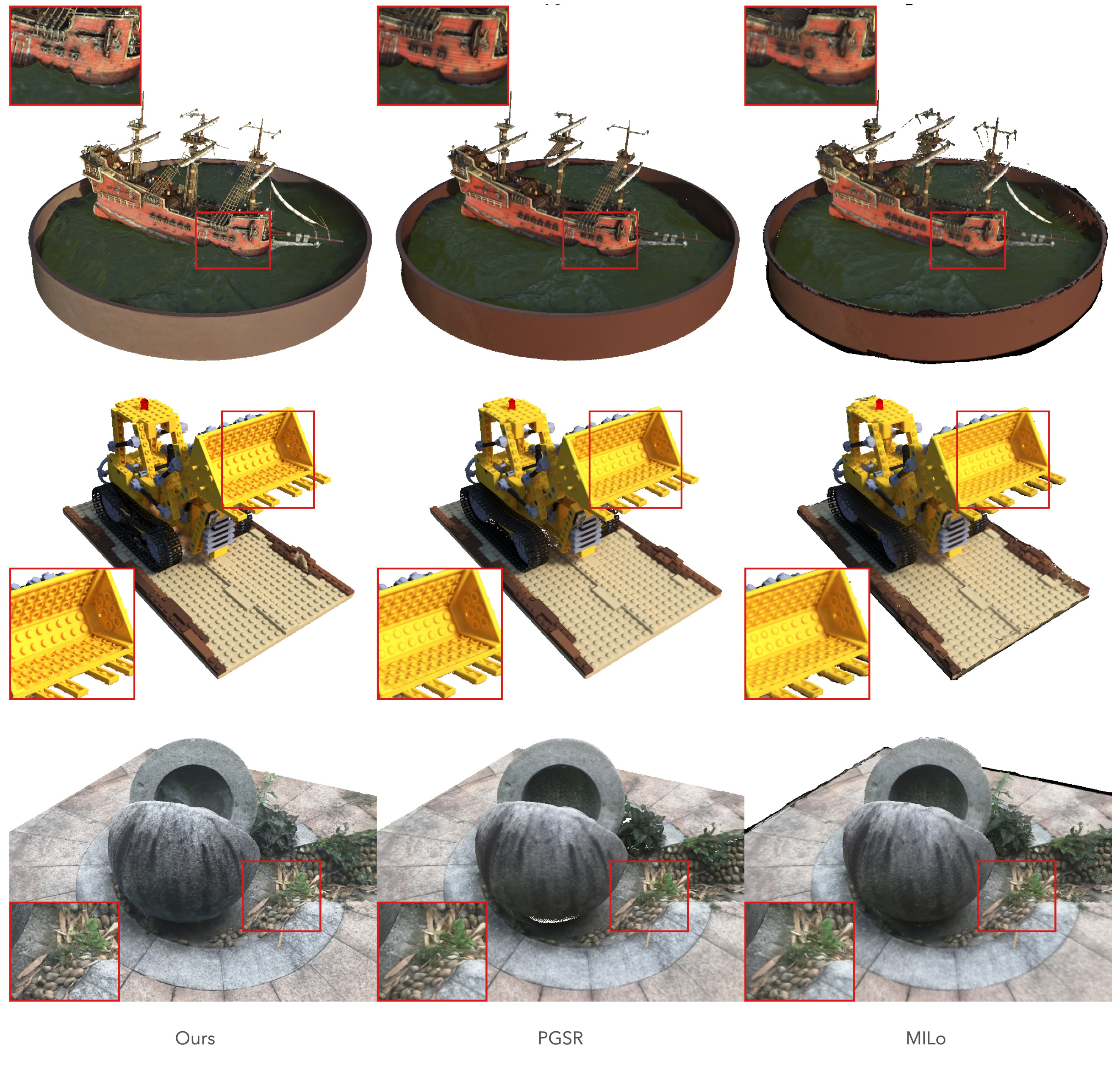}
    \caption{
    \textbf{Diffuse-shaded renderings of extracted geometry.}
    Comparison on representative scenes from NeRF Synthetic and BlendedMVS. ADELE produces sharp, realistic renderings that preserve fine appearance detail, whereas PGSR and MILo smooth over high-frequency content as a consequence of their less accurate geometric reconstruction.
   }\label{fig:nvs_results}
\end{figure*}

We report additional novel view synthesis results on DTU and Stanford ORB for ADELE, MILo, and PGSR (\autoref{tab:nvs_results}). For the ORB dataset these correspond to PSNR on unseen views, whereas for DTU we report PSNR for the training views as there is no common test/training split and the methods were trained on all images. Figure~\ref{fig:nvs_results} shows a comparison of diffuse renderings of the extracted meshes.

Under each method's \emph{native} rendering, PGSR and MILo benefit from their full volumetric 3DGS pipeline, while our method renders a single-surface intersection via mesh rasterization. This gap in expressiveness is reflected in the average PSNR scores. Under the diffuse shading protocol ADELE outperforms PGSR and MILo on both Stanford ORB and DTU. Qualitatively (Fig.~\ref{fig:nvs_results}), our diffuse renderings preserve sharp, high-frequency appearance detail, whereas PGSR and MILo smooth over fine structure as a consequence of less accurate underlying geometry. 
We acknowledge that the extracted-mesh rendering protocol is also an imperfect appearance metric. By reducing each reconstruction to a diffuse-shaded mesh, it discards the view-dependent appearance the underlying methods model, so lower absolute PSNR values under this protocol do not necessarily indicate worse geometry. We nevertheless report it as a complementary evaluation to the native-rendering scores, which structurally favor volumetric and Gaussian-based methods.

\begin{table*}[h]
\caption{Novel-view synthesis under two rendering protocols. \emph{Native rendering} uses each method's own renderer (3DGS for PGSR and MiLo, neural-shaded mesh rasterization for ours); \emph{Extracted-mesh rendering} renders the extracted mesh shaded with a per-vertex diffuse albedo. We report PSNR ($\uparrow$) on Stanford ORB (top) and DTU (bottom).}
\label{tab:nvs_results}
\centering
% ---- Stanford ORB ----
\begin{tabular}{ll|cccccc}
\multicolumn{2}{c}{} & \multicolumn{6}{c}{PSNR$\uparrow$ -- Stanford ORB} \\
Render & Method & Gnome & Cactus & Ball & Teapot & Pitcher & Avg \\
\hline
\multirow{3}{*}{Native}
       & PGSR  & \textbf{37.91} & \textbf{37.85} & 36.91 & 37.43 & 31.46 & \textbf{36.31} \\
       & MiLo  & 37.01 & 37.17 & 34.80 & \textbf{37.57} & \textbf{32.98} & 35.91 \\
       & Ours  & 35.24 & 35.09 & \textbf{37.45} & 35.85 & 32.96 & 35.31 \\
\hline
\multirow{3}{*}{Mesh}
       & PGSR  & 30.93 & \textbf{32.25} & \textbf{31.04} & 31.56 & 27.91 & 30.74 \\
       & MiLo  & 30.34 & 31.73 & 29.38 & 29.71 & \textbf{29.56} & 30.14 \\
       & Ours  & \textbf{31.27} & 31.73 & 30.80 & \textbf{32.15} & 28.33 & \textbf{30.86} \\
\hline
\end{tabular}

\vspace{1.5em}

% ---- DTU ----
\resizebox{\linewidth}{!}{%
\begin{tabular}{ll|cccccccccccccccc}
\multicolumn{2}{c}{} & \multicolumn{16}{c}{PSNR$\uparrow$ -- DTU} \\
Render & Method & 24 & 37 & 40 & 55 & 63 & 65 & 69 & 83 & 97 & 105 & 106 & 110 & 114 & 118 & 122 & Avg \\
\hline
\multirow{3}{*}{Native}
       & PGSR & 34.62 & 30.39 & 34.12 & 36.59 & 39.02 & 36.21 & 34.13 & \textbf{42.95} & 34.60 & \textbf{39.49} & 38.45 & 37.05 & 33.82 & 40.68 & 41.66 & 36.92 \\
       & MiLo & \textbf{35.77} & \textbf{32.94} & \textbf{35.43} & \textbf{37.24} & \textbf{41.41} & \textbf{37.35} & \textbf{35.40} & 42.59 & \textbf{34.82} & 38.90 & \textbf{41.03} & \textbf{38.06} & \textbf{34.69} & \textbf{42.18} & \textbf{42.95} & \textbf{38.05} \\
       & Ours & 30.99 & 28.67 & 30.98 & 33.70 & 35.40 & 34.13 & 31.49 & 36.88 & 30.84 & 34.42 & 36.95 & 36.08 & 32.40 & 38.91 & 40.59 & 34.16 \\
\hline
\multirow{3}{*}{Mesh}
       & PGSR & 14.36 & 13.84 & 16.11 & 17.32 & 19.22 & 14.90 & \textbf{19.32} & 21.83 & 17.87 & 18.08 & 20.50 & 19.44 & 16.29 & 20.89 & \textbf{21.58} & 18.10 \\
       & MiLo & 14.30 & 13.68 & 15.93 & 17.05 & 19.10 & 14.71 & 19.04 & 21.57 & 17.60 & 17.82 & 20.32 & 19.31 & 16.07 & 20.77 & 21.48 & 17.92 \\
       & Ours & \textbf{14.44} & \textbf{13.96} & \textbf{16.37} & \textbf{17.51} & \textbf{19.47} & \textbf{15.36} & 18.95 & \textbf{22.00} & \textbf{18.33} & \textbf{18.55} & \textbf{20.89} & \textbf{19.74} & \textbf{16.61} & \textbf{20.91} & 21.28 & \textbf{18.29} \\
\hline
\end{tabular}
}
\end{table*}

\subsection{Qualitative Results}

\begin{figure*}[t]
  \centering
  \includegraphics[width=1.0\linewidth]{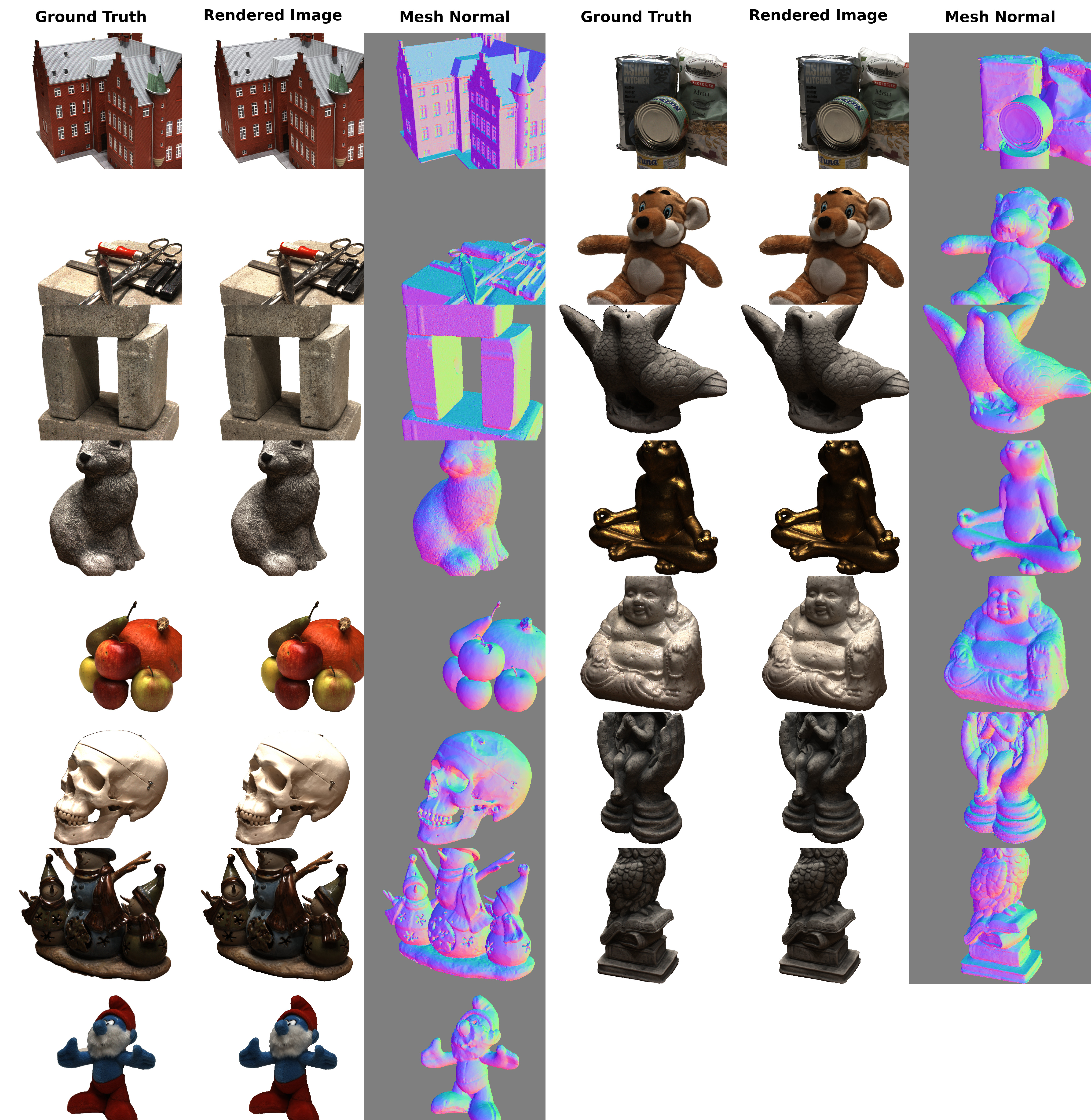}
    \caption{Full results of our method on the 15 scenes of the DTU dataset used in the evaluation of the main paper.}
   \label{fig:additional_dtu_results}
\end{figure*}

\begin{figure*}[t]
  \centering
  \includegraphics[width=1.0\linewidth]{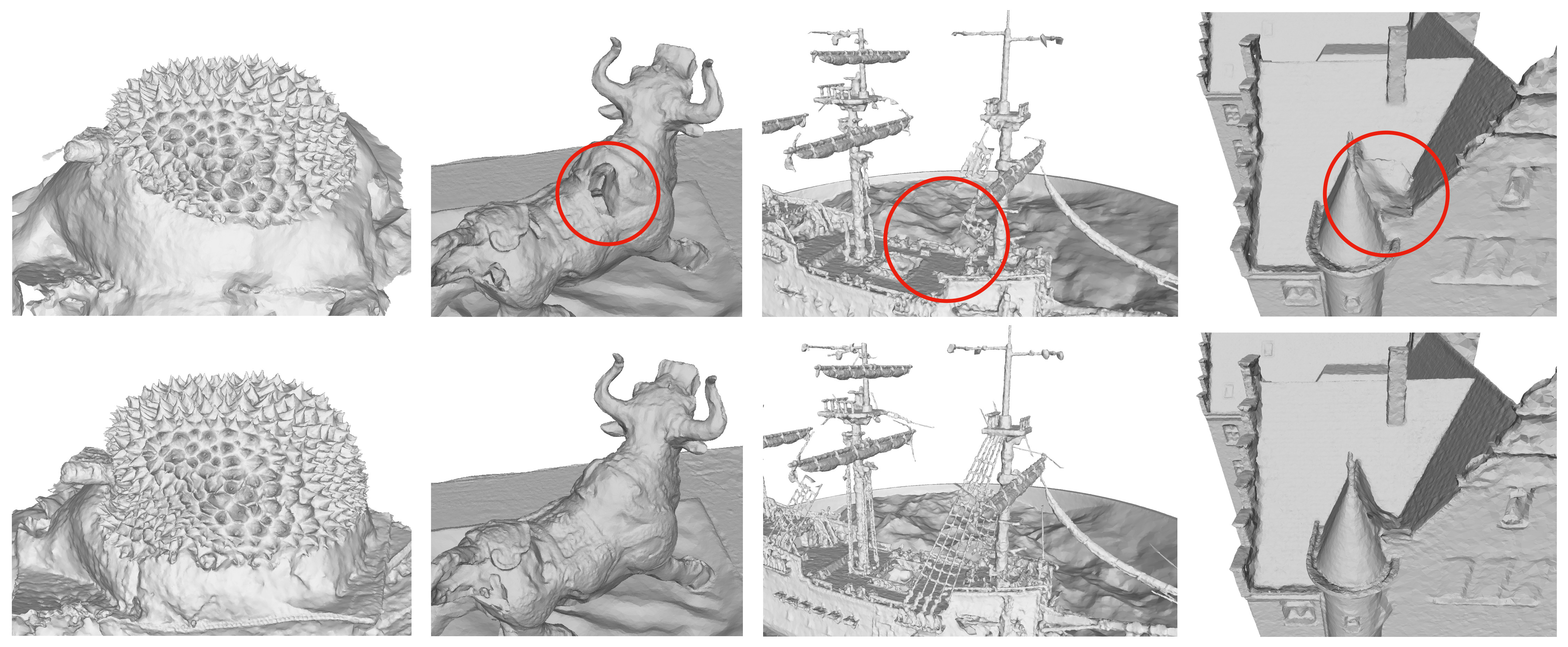}
    \caption{
    \rev{Effect of depth offset rendering. Training with regular mesh
  rendering (top) produces both convex and concave surface artefacts across a
  variety of scenes. Training with depth offset rendering (bottom) largely
  eliminates them.}}
   \label{fig:offset_rendering_ablation}
\end{figure*}

\autoref{fig:additional_dtu_results} presents normals of meshes reconstructed with our method for all scenes of the DTU dataset used in the evaluation. \rev{\autoref{fig:offset_rendering_ablation} provides additional ablation
results for the depth offset rendering technique, illustrating how it
alleviates artefacts that arise when training with regular
mesh-based rendering.}

\end{document}